%% file: main.tex
\documentclass{article}

\usepackage{microtype}
\usepackage{graphicx}
\usepackage{booktabs} 
\usepackage{environ}%
\usepackage[CJKbookmarks=true,
            bookmarksnumbered=true,
            bookmarksopen=true,
            colorlinks=true,
            citecolor=blue,
            linkcolor=red,
            anchorcolor=red,
            urlcolor=blue
            ]{hyperref}
\usepackage{geometry}
\usepackage{algorithm}
\usepackage{algorithmic}

\usepackage{amsmath}
\usepackage{amssymb}
\usepackage{mathtools}
\usepackage{amsthm}

\usepackage[capitalize,noabbrev]{cleveref}

\theoremstyle{plain}
\newtheorem{theorem}{Theorem}[section]

\newtheorem{lemma}[theorem]{Lemma}

\theoremstyle{definition}

\newtheorem{assumption}[theorem]{Assumption}
\theoremstyle{remark}
\newtheorem{remark}[theorem]{Remark}

\usepackage[textsize=tiny]{todonotes}

\title{Principled Reinforcement Learning with Human Feedback from Pairwise or $K$-wise Comparisons}
\usepackage{babel}
\usepackage{verbatim}
\usepackage{mathtools}
\usepackage{bm}
\usepackage{natbib}
\usepackage{amsmath}
\usepackage{amssymb}
\usepackage{multicol}
\setcitestyle{authoryear,open={(},close={)}} 

\makeatletter

\newcommand{\upstairs}[1]{\textsuperscript{#1}}
\newcommand{\affilone}{\dag}
\newcommand{\affiltwo}{\ddag}

\input{macro.tex}

\input{macro2.tex}

\usepackage{bm}
\usepackage{enumitem}

\usepackage{babel}

\usepackage{amsthm}
\usepackage{bbm}

\theoremstyle{plain}

\usepackage{caption}
\usepackage{subcaption}
\theoremstyle{definition}

\usepackage{color}
\definecolor{cm}{RGB}{0,0,200}
\definecolor{purple}{RGB}{200,0,200}

\NewEnviron{resizealign}{\sbox0{
    $\begin{matrix}\displaystyle\BODY\end{matrix}$}%
  \sbox1{$(\theequation)$}%
  \sbox2{\parbox{\dimexpr \wd0 + 2\wd1}%
    {\begin{align}\BODY\end{align}}}
  \noindent\resizebox{\hsize}{!}{\usebox2}%
}
\begin{document}

\begin{center}

  {\bf{\Large Principled Reinforcement Learning with Human Feedback from Pairwise or $K$-wise Comparisons}} \\

  \vspace*{.2in}
  
  \begin{tabular}{cccc}
  
    Banghua Zhu\upstairs{\affilone}\quad

    Michael I.\ Jordan\upstairs{\affilone, \affiltwo} \quad
 
        Jiantao Jiao\upstairs{\affilone, \affiltwo}
    \vspace*{.1in} \\

    \upstairs{\affilone} Department of Electrical Engineering and Computer Sciences, UC Berkeley \\
    \upstairs{\affiltwo} Department of Statistics, UC Berkeley
  \end{tabular}
  
  \vspace*{.2in}
\end{center}
\begin{abstract}
    We  provide a theoretical framework for Reinforcement Learning with Human Feedback (RLHF).  Our analysis shows that when the true reward function is linear, the widely used maximum likelihood estimator (MLE) converges under  both the Bradley-Terry-Luce (BTL) model and the Plackett-Luce (PL) model. However, we show that when training a policy based on the learned reward model, MLE fails while a pessimistic MLE provides policies with improved performance under certain coverage assumptions. Additionally, we demonstrate that under the PL model, the true MLE and an alternative MLE that splits the $K$-wise comparison into pairwise comparisons both converge. Moreover, the true MLE is asymptotically more efficient. Our results validate the empirical success of existing RLHF algorithms in InstructGPT and provide new insights for algorithm design. We also unify the problem of RLHF and max-entropy Inverse Reinforcement Learning (IRL), and provide the first sample complexity bound for  max-entropy IRL. 
\end{abstract}

\section{Introduction}

The alignment problem aims at aligning human values with machine learning systems and steering learning algorithms towards the goals and interests of humans.  One of the most promising tools for AI alignment, 
\emph{Reinforcement Learning with Human Feedback} (RLHF, or Preference-based Reinforcement Learning), has delivered significant empirical success in the fields of game playing, robot training, stock-prediction, recommender systems, clinical trials, large language models  etc. \citep{sui19,sadigh17,nipsPRL17,kupcsik18,jain13,wirth17, knox2008tamer, macglashan2017interactive, christiano2017deep, warnell2018deep, brown2019extrapolating, shin2023benchmarks, ziegler2019fine, stiennon2020learning, wu2021recursively, nakano2021webgpt,ouyang2022training, menick2022teaching, glaese2022improving, gao2022scaling, bai2022training, ganguli2022red, ramamurthy2022reinforcement}. Notably, the language model application ChatGPT is based on RLHF and this underlies several of its skills: answering followup questions, admitting its mistakes, challenging incorrect premises, and rejecting inappropriate requests. One of the key capabilities in RLHF is to learn a reward from human feedback,  in the form of pairwise or $K$-wise comparisons between actions (responses).  In this paper, we take the first step towards providing  a theoretical framework for  RLHF, with a specific focus on reward learning. We  provide theoretical analysis that  justifies the empirical success of RLHF in InstructGPT and ChatGPT, along with new insights for  algorithm design.

Taking InstructGPT~\cite{ouyang2022training} as an example, a typical deployment of RLHF for language modeling includes the following steps:

(a) Pre-train a Large Language Model (LLM) using supervised training.

(b) Train a reward model based on the pre-trained LLM using human feedback. 

 (c) Fine-tune the existing LLM based on the learned reward model using Proximal Policy Optimization (PPO).

During the reward training step, the prompts are  first sampled   from a pre-collected dataset. Then  $K$ responses are sampled by executing existing models on the sampled prompts. Based on the prompt provided, a human labeler ranks all the responses according to  her own preference. The reward model is  trained based on a maximum likelihood estimator (MLE), also known as the learning-to-rank algorithm or cross-entropy minimization~\citep{liu2009learning, xia2008listwise, cao2007learning, christiano2017deep, ouyang2022training}.

In the setting of InstructGPT, the ranking of responses is  based purely on the current prompt, which can be viewed as the state in a contextual bandit. We accordingly start with the setting of a contextual bandit, and later generalize our results to  Markov Decision Process (MDP) where there are transitions between states. Let $\mathcal{S}$ be the set of states (prompts), and $\mathcal{A}$ be the set of actions (responses). For each state-action pair $(s, a)$, we assume that the reward is parametrized by $r_\theta(s, a) = \inprod{\theta}{\phi(s, a)}$ for some  known and fixed feature function $\phi(s, a):\mathcal{S}\times\mathcal{A}\mapsto \mathbb{R}^d$. In an LLM, such a $\phi$ is usually derived by removing the last layer of the pre-trained model.\footnote{In InstructGPT, the function $\phi$ is still parametrized can be further trained in the reward learning step. However, for simplicity of theoretical analysis we assume in this paper that $\phi$ is fixed and one only fine-tunes the last layer with parameter $\theta$.} We denote the ground-truth reward provided by a human as $r_{\theta^\star}(s, a)$ for some parameter $\theta^\star\in\mathbb{R}^d$.

We are interested in the sample complexity for learning a reward model $r_{\theta^\star}$ from pairwise or $K$-wise comparison data.  For the $i$-th sample, a state  $s^i$ is first sampled from some fixed distribution $\rho$. Given the state $s^i$, $K$ actions $(a_0^i, a_1^i,\cdots, a_{K-1}^i)$ are sampled from some joint distribution $\mathbb{P}(a_0,\cdots, a_{K-1} \mid s^i)$. Let $\sigma^i:[K]\mapsto [K]$ denote the output of the human labeller, which is a permutation function representing the ranking of  the actions. Here $\sigma^i(0)$ represents the most preferred action. We assume that the distribution of $\sigma^i$ follows a Plackett-Luce (PL) model~\citep{plackett1975analysis,luce2012individual}:
\begin{align}
\mathbb{P}(\sigma^i \mid s^i, a_0^i, a_1^i,\cdots, a^i_{K-1}) = \prod_{k=0}^{K-1} \frac{\exp(r_{\theta^\star} (s^i, a^i_{\sigma^i(k)}))}{\sum_{j=k}^{K-1}\exp(r_{\theta^\star} (s^i, a^i_{\sigma^i(j)}))}. \nonumber
\end{align}
When $K=2$, this reduces to the pairwise comparison of the Bradley-Terry-Luce (BTL) model~\citep{bradley1952rank}, which is widely applied  in  existing RLHF algorithms~\cite{christiano2017deep, ouyang2022training}. 

Since the learned reward model is mainly used for downstream policy training, we measure the correctness of the estimated reward model via the performance of a greedy policy trained from a reward model $r_{\hat\theta}$. Concretely, for a greedy policy $\hat\pi(s) = \argmax_a r_{\hat \theta}(s, a)$, we compute a performance gap compared to the optimal policy:
\begin{align*}
\mathsf{SubOpt}(\hat\pi) \coloneqq
\mathbb{E}_{s\sim\rho} [r_{\theta^\star}(s, \pi^\star(s)) -r_{\theta^\star}(s, \hat \pi(s)].
\end{align*}
Here  $\pi^\star=\argmax_a r_{\theta^\star}(s, a)$ is the optimal policy under the true reward $r_{\theta^\star}$.

\citet{gao2022scaling} has observed that in the  reward model trained from practice, there exists an over-optimization phenomenon where the true reward first increases and then decreases during the policy optimization stage.  In this paper, we study the potential sub-optimality of the  MLE in the RLHF setting. As a by-product, we also provide guarantee of the estimation error on the   semi-norm of the parameter estimation error, $\|\hat \theta - \theta^\star\|_{\Sigma}$, for a query-dependent covariance matrix $\Sigma$.

From a broader perspective, the framework of  RLHF can be viewed as a special case of reward learning from pre-collected data, which has been a primary focus in Inverse Reinforcement Learning (IRL) and offline reinforcement learning. Our techniques also provide theoretical guarantee for the  max-entropy IRL~\cite{ziebart2008maximum}  and action-based IRL  algorithms~\cite{ramachandran2007bayesian, neu2009training, florence2022implicit}. 

\subsection{Main Results}
\paragraph{Pairwise Comparison.}
We start with the setting of a contextual bandit with pairwise comparison.
We focus on two algorithms, MLE and pessimistic MLE. The following result from  dueling bandits and RL~\citet{faury2020improved, pacchiano2021dueling}  shows that under a semi-norm $\|\cdot\|_\Sigma$, MLE converges to the true parameter.  
\begin{lemma}[Informal]\label{thm:informal_MLE_pair}
Under certain regularity conditions, the MLE satisfies the following with probability at least $1-\delta$, 
   \begin{align*}
\|\hat\theta_{\mathsf{MLE}} - \theta^\star\|_{\Sigma_{\mathcal{D}}}\leq C\cdot \sqrt{ \frac{d+\log(1/\delta)}{n}}.
   \end{align*} 
   Here $\Sigma_\mathcal{D} = \frac{1}{n}\sum_{i=1}^n (\phi(s^i, a_1^i) - \phi(s^i, a_0^i))(\phi(s^i, a_1^i) - \phi(s^i, a_0^i))^\top $.
\end{lemma} 
However, when we consider the performance of the induced policy, MLE provably fails  while pessimistic MLE gives a near-optimal rate. In essence, the pessimism principle discounts actions that are less represented in the observed dataset, and hence is conservative in outputting a policy. 
\begin{theorem}[Informal]
Under certain coverage assumption, one can design a pessimistic MLE such that the induced greedy policy $\hat \pi_{\mathsf{PE}}$  is good; i.e., with probability at least $1-\delta$,
\begin{align*}
    \mathsf{SubOpt}(\hat\pi_{\mathsf{PE}}) = \Theta\left( \sqrt{\frac{d+\log(1/\delta)} {n}}\right).
\end{align*}
In contrast, under the same assumption, one can find instances such that the greedy policy w.r.t.\ MLE  $\hat\pi_{\mathsf{MLE}}$ fails:
\begin{align*}
   \forall n>1, \mathbb{E}[\mathsf{SubOpt}(\hat\pi_{\mathsf{MLE}})] \geq 0.1.
\end{align*}
\end{theorem}

\paragraph{$K$-wise Comparison.}
For $K$-wise comparison, we analyze both the MLE and the algorithm in InstructGPT~\citep{ouyang2022training} which splits the ranking data into $K(K-1)/2$ pairwise comparison data and runs an MLE based on the BTL model. We show that both converge in terms of the estimation error under the semi-norm, and give a near-optimal policy when combined with pessimism. More importantly, we show that although both estimators are unbiased, the asymptotic variance of MLE is smaller than that of the splitted estimator in  InstructGPT~\citep{ouyang2022training}, which belongs to the family of M-estimators. Thus the MLE is more efficient than the existing algorithm used in InstructGPT. We also conduct experiments to verify the theoretical prediction. 

Let the estimated parameter for the splitted estimator be $\hat\theta$ and the induced policy be $\hat \pi_{\mathsf{PE}}$. We have:
\begin{theorem}[Informal]
Under certain coverage and  regularity conditions, the following holds separately with probability at least $1-\delta$:
   \begin{align*}
   \|\hat\theta-\theta^\star\|_{\Sigma_{\mathcal{D}}}&\leq  C \cdot  \sqrt{ \frac{d+\log(1/\delta)}{n}}, \\ 
   \mathsf{SubOpt}(\hat\pi_{\mathsf{PE}}) &\leq C' \cdot \sqrt{ \frac{d+\log(1/\delta)}{n}}, 
   \end{align*} 
   Here $\Sigma_{\mathcal{D}} =  \frac{2}{K(K-1)\numobs}  (\sum_{i=1}^{\numobs} 
\sum_{j=0}^{\numchoices-1} \sum_{k=j+1}^{K-1} (\phi(s^i,a^i_{j}) - \phi(s^i,a^i_{k}))(\phi(s^i,a^i_{j}) - \phi(s^i,a^i_{k}))^\top). $
\end{theorem}

We also extend our  results to the case of MDP and IRL; see the detailed presentation in Section~\ref{sec:MDP} and Section~\ref{sec:irl}. Let the estimated parameter be $\hat\theta$ and the induced pessimistic policy be $\hat \pi_{\mathsf{PE}}$. For pairwise comparison we have:
\begin{theorem}[Informal]
In the MDP setting with horizon $H$, under certain coverage and  regularity conditions, the following holds separately with probability at least $1-\delta$:
   \begin{align*}
   \|\hat\theta-\theta^\star\|_{\Sigma_{\mathcal{D}}}&\leq  C \cdot \sqrt{ \frac{d+\log(1/\delta)}{n}}, \\ 
   \mathsf{SubOpt}(\hat\pi_{\mathsf{PE}}) &\leq C'  \cdot \sqrt{ \frac{d+\log(1/\delta)}{n}}, 
    \end{align*} 
   Here $\Sigma_{\mathcal{D}} = \frac{1}{n}\sum_{i=1}^n (\sum_{h=0}^H (\phi(s^i_h, a_{h}^i)-\phi(s_h^{i\prime}, a_{h}^{i\prime})))$   $(\sum_{h=0}^H (\phi(s^i_h, a_{h}^i)-\phi(s_h^{i\prime}, a_{h}^{i\prime})))^\top$.
\end{theorem}

Our results not only explain the correctness of existing algorithms, but also provide new insights for algorithm design in RLHF.  In particular, it suggests the importance of introducing pessimism in the reward learning part, which can be implemented via adding regularization in  policy training steps as in~\citet{ouyang2022training}, or using existing offline RL algorithms, including but not limited to Conservative Q-Learning~\citep{kumar2020conservative}, Implicit Q-Learning~\citep{kostrikov2021offline} and Adversarially Trained Actor Critic~\citep{cheng2022adversarially}. On the other hand, it also shows that MLE is a more efficient estimator than that in~\cite{ouyang2022training}.  

\subsection{Related Work}

\paragraph{Learning and Estimation from Pairwise Comparison and Ranking.} 
The problem of estimation and ranking from pairwise or $K$-wise comparisons has been studied extensively in the literature. In the literature of \textit{dueling bandit},  one compares two actions and aims to minimize regret based on pairwise comparisons~\citep{Yue+12,Zoghi+14RCS, Yue+09,BTM,RDB,GS21,SG18, Ailon+14,Zoghi+14RUCB,Komiyama+15,Adv_DB,SGrank18,SGwin18, faury2020improved}.
\cite{sui19, xu20}  analyze the sample complexity of dueling RL under the tabular case, which is extended to linear case and function approximation by the recent work~\citet{pacchiano2021dueling, chen2022human}. \citet{chatterji2022theory} studies a close setting where in each episode only binary feedback is received. However, most of the work focuses on  regret minimization. We take a first step towards the theoretical analysis for function approximation for $K$-wise comparisons with policy learning as the target.

On the other hand, in the literature of ranking,  most of the theoretical work focuses on the tabular case where the rewards for different actions are uncorrelated~\citep{feige1994computing, shah2015estimation, shah2017simple, heckel2018approximate, mao2018minimax, jang2017optimal, chen2013pairwise, chen2015spectral, rajkumar2014statistical,negahban2018learning, hajek2014minimax, heckel2019active}. And a majority of the empirical literature focuses on the framework of learning to rank (MLE) under general function approximation, especially when the reward is parameterized by a neural network~\citep{liu2009learning, xia2008listwise, cao2007learning, christiano2017deep, ouyang2022training, brown2019extrapolating, shin2023benchmarks, busa14,wirth16,wirth17,nipsPRL17, abdelkareem2022advances}. Similar idea of RL with AI feedback also learns a reward model from preference~\cite{bai2022constitutional}, except for that  the preference is labeled by another  AI model instead of human.

\paragraph{Inverse Reinforcement Learning and Offline Reinforcement Learning.}

RLHF, IRL and offline learning are all approaches that can be used to incorporate human preferences or expertise into the decision-making process of an  agent. However, they differ in the way that they use human input to guide the agent's behavior.   In IRL and imitation learning, we only observe  an expert's behavior and would like to infer the expert's preferences or goals~\citep{ng2000algorithms, abbeel2004apprenticeship, ziebart2008maximum,  ramachandran2007bayesian, neu2009training, ho2016generative, florence2022implicit, hussein2017imitation}. In offline learning, we directly observe the cardinal rewards for the state. But the actions are likely to be sub-optimal. In RLHF, we observe ordinal comparisons between pairs or a set of actions. 
In one of the popular IRL frameworks, max-entropy IRL~\citep{ziebart2008maximum}, it is also assumed that human choice follows a PL model.  We unify the problem of RLHF and max-entropy IRL, and provide the first sample complexity analysis for max-entropy IRL. 

\paragraph{Pessimism in Offline RL. }
The idea of introducing pessimism for offline RL has been  studied in recent year~\citep{jin2021pessimism, rashidinejad2021bridging, li2022pessimism, xie2021policy, zanette2022realizability, zanette2021provable, xie2021bellman, xu2022provably}.  In this paper, we connect RLHF with offline RL and show that pessimism also helps in RLHF. 
\section{Preliminaries}

We begin with the notation that we use in the paper. We  discuss our formulations of contextual bandits and  Markov decision processes in Section~\ref{sec:MDP_model}.
We introduce the data collection model and the BTL and PL models in Section~\ref{sec:pre-sampling}.

\textbf{Notations.} We use calligraphic letters for sets, e.g., $\mathcal{S}$ and $ \mathcal{A}$. Given a set $\mathcal{S}$, we write $|\mathcal{S}|$ to represent the cardinality of $\mathcal{S}$.  For    vectors $x$ and $y$, we use $\inprod{x}{y} = x^\top y$ to denote
their inner product. 
We use $[K]$ to denote the set of integers from $0$ to $K-1$. 
We write $\|x\|_\Sigma = \sqrt{x^\top \Sigma x}$
 as a semi-norm of $x$ when $\Sigma$ is some positive-semidefinite matrix.  We write $\Sigma\succeq \Sigma'$ if $\Sigma-\Sigma'$ is  positive semidefinite.
 
 \subsection{Markov decision processes} \label{sec:MDP_model}
 
 We consider a finite-horizon  MDP described by a tuple $M = (\mathcal{S}, \mathcal{A}, H, \{P_h\}_{h=1}^H, \{R_h\}_{h=1}^H, \rho)$, where $\mathcal{S}$ is a (possibly infinite) state space, $\mathcal{A}$ is a (possibly infinite) action space, $H$ is the horizon length, $P_h: \mathcal{S} \times \mathcal{A} \mapsto \Delta(\mathcal{S})$ is a probability transition matrix at step $h$, $R_h: \mathcal{S} \times \mathcal{A} \mapsto \Delta([0,1])$ encodes a family of reward distributions with $r_h: \mathcal{S} \times \mathcal{A} \mapsto [0,1]$ as the expected reward function, $\rho: \mathcal{S} \mapsto \Delta(\mathcal{S})$ is the initial state distribution. At step $h$, upon executing action $a$ from state $s$, the agent receives a deterministic reward   $r_h(s,a)$ and transits to the next state $s'$ with probability $P_h(s'|s,a)$. The MDP transits to an absorbing termination state with zero reward at step $H$. When $H=1$ and there is no transition, the model reduces to the contextual bandit problem.

A  deterministic policy $\pi_h: \mathcal{S} \mapsto \mathcal{A}$ is a function that maps a state to an action at step $h\in[H]$. We use $\pi$ to denote the family of policies $\{\pi_h\}_{h=1}^H$.
Correspondingly, the value function $V^\pi: \mathcal{S} \mapsto \mathbb{R}$ of the policy family $\{\pi_h\}_{h\in[H]}$ is defined as the expected sum of  rewards starting at state $s$ and following policy $\pi_h$ at step $h$. More precisely, we have for any $s\in\mathcal{S}$,
$  V^\pi(s) \coloneqq \E \left[\sum_{h=0}^H  r_h(s_h, a_h) \mid s_0 = s , a_h = \pi_h(s_h),\forall  h\geq 0\right]$,
where the expectation is taken over the trajectory generated according to the transition kernel $s_{h+1} \sim P_h(\cdot \mid s_h, a_h)$ and reward distribution $r_h \sim R_h(\cdot \mid s_h, a_h)$. The Q-function  $Q^\pi: \mathcal{S} \times \mathcal{A} \rightarrow \mathbb{R}$ of policy $\pi$ is defined analogously: $
    Q^\pi(s,a) \coloneqq \E \left[\sum_{h=0}^H r_h(s_h,a_h) \;\mid\; s_0 = s, a_0 = a, a_h = \pi_h(s_h), \forall h\geq 0 \right].$ Note that although we work with undiscounted episodic case, it is straightforward to extend the framework and analysis to discounted MDP.
 We   define the expected value of a policy $\pi$:
\begin{align*} 
    J(\pi) \coloneqq \E_{s\sim \rho}[V^\pi(s)] = \sum_{s \in \mathcal{S}} \rho(s)V^\pi(s).
\end{align*}
We use shorthands $V^\star \coloneqq V^{\pi^\star}$ and $Q^\star \coloneqq Q^{\pi^\star}$ to denote the optimal value function and the optimal Q-function. We define the sub-optimality of any policy $\pi$ as 
\begin{align*}
\mathsf{SubOpt}(\pi) \coloneqq
J(\pi^\star) - J(\hat \pi) .
\end{align*}
We use shorthands $V^\star \coloneqq V^{\pi^\star}$ and $Q^\star \coloneqq Q^{\pi^\star}$ to denote the optimal value function and the optimal Q-function. We define the sub-optimality of any policy $\pi$ as 
\begin{align*}
\mathsf{SubOpt}(\pi) \coloneqq
J(\pi^\star) - J(\hat \pi) .
\end{align*}
We also define the state  occupancy measures $d^\pi: \mathcal{S} \mapsto [0,H]$ and state-action occupancy measures $d^\pi: \mathcal{S}\times \mathcal{A} \mapsto [0,H]$   as
$
    d^\pi(s)  \coloneqq   \sum_{h=0}^H \mathbb{P}_h(s_h = s \mid  \pi),  
    d^\pi(s,a) \coloneqq  \sum_{h=0}^H \mathbb{P}_h(s_h = s; a_h=a \mid \pi),$ 
where we use  $\mathbb{P}_h(s_h = s \mid  \pi)$ to denote the probability of visiting state $s_h = s$ (and similarly $s_h = s, a_h = a$) at step $h$ after executing policy $\pi$ and starting from $s_0\sim \rho(\cdot)$.

Throughout the paper, we make the following assumption on the parameterization of the reward:
\begin{assumption}
The reward lies in the family of linear functions  $r_\theta(s, a) = \theta^\top \phi(s, a)$ for some known $\phi(s, a)$ with $\max_{s, a}\|\phi(s, a)\|_2\leq L$. Let $\theta^\star$ be the true parameter. To ensure the identifiability of $\theta^\star$, we let $\theta^\star\in \Theta_B$, where 
\begin{align*}
    \Theta_B = \{\theta\in\mathbb{R}^d \mid \inprod{1}{\theta}=0, \|\theta\|_2\leq B\}.
\end{align*}
\end{assumption}
\subsection{Sampling Procedure and Comparison Model}\label{sec:pre-sampling} 
As in~\citet{ouyang2022training},
we assume that both the states and actions in the training set come from a pre-collected dataset.
In a contextual bandit, 
for the $i$-th  sample, a state (prompt) $s^i$ is first sampled from some fixed distribution $\rho$. Given the state $s^i$, $K$ actions $(a_0^i, a_1^i,\cdots, a_{K-1}^i)$ are sampled from some joint distribution $\mathbb{P}(a_0,\cdots, a_{K-1} \mid s^i)$\footnote{Indeed, it is not necessary to only compare actions under the same state. Our results can be easily generalized to the case when the states for $K$ queries are completely different.}. Let $\sigma^i:[K]\mapsto [K]$ be the output of the human labeller, which is a permutation function that denotes the ranking of  the actions. Here $\sigma^i(0)$ represents the most preferred action. We  use $a_0>a_1$ to denote the event that the action $a_0$ is more preferred compared to $a_1$.  
A common model on the distribution of $\sigma$ under $\numchoices$-ary     comparisons is  a Plackett-Luce model~\citep{plackett1975analysis,luce2012individual}.
The Plackett-Luce model  defines the probability of  a state-action pair $(s, a_i)$ being the largest among a given set $\{(s, a_i)\}_{i=0}^{K-1}$ as
\begin{align*}
\mathbb{P}(a_i > a_j, \forall j\neq i \mid s)=
\frac{\exp(r_{\theta}{(s, a_i)}) }{\sum_{j=0}^{\numchoices-1} \exp(r_{\theta}{(s, a_j)})}. 
\end{align*}
Moreover, one can calculate the probability of observing the permutation $\sigma$ as\footnote{In practice, one may introduce an extra temperature parameter $\sigma$ and replace all $r_{\theta^\star}$ with $r_{\theta^\star}/\sigma$. Here  we take  $\sigma=1$.}
\begin{align*}
\mathbb{P}(\sigma\mid s, \{a_i\}_{i=0}^{K-1})=
\prod_{i=0}^{K-1}\frac{\exp(r_{\theta^\star}(s, a_{\sigma(i)}) )}{\sum_{j=i}^{\numchoices-1} \exp(r_{\theta^\star}(s, a_{\sigma(j)}))}.
\end{align*}

When $K=2$, this reduces to the pairwise comparison considered in the BTL model, which is  used in existing RLHF algorithms. In this case, the permutation $\sigma$ can be reduced to a Bernoulli random variable, representing whether $a_0$ is preferred compared to $a_1$. Concretely, for each queried state-actions pair $(s, a_0, a_1)$, we  observe a sample $y$ from a Bernoulli distribution with parameter $\frac{\exp(r_{\theta^\star}(s, a_{1}))}{\exp(r_{\theta^\star}(s, a_{0}))+\exp(r_{\theta^\star}(s, a_{1}))}$; i.e., for any $l\in\{0, 1\}$,
\begin{align*}
    \mathbb{P}(y = l \mid s, a_0, a_1)& = \frac{\exp(r_{\theta^\star}(s, a_{l}))}{\exp(r_{\theta^\star}(s, a_{0}))+\exp(r_{\theta^\star}(s, a_{1}))}. 
\end{align*}

\subsection{Organization}
Section~\ref{sec:pairwise} presents the problem of learning with pairwise comparisons under the contextual bandit framework, we provide upper and lower bounds for MLE and pessimistic MLE. We extend the result into K-wise comparisons in Section~\ref{Sec:Kwise} and MDP in Section~\ref{sec:MDP}. We discuss the guarantee for IRL in Section~\ref{sec:irl}. We present our experimental results on simulated dataset in Section~\ref{app:experiments}. We also discuss the analysis for nonlinear rewards in Appendix~\ref{app:nonlinear} .

\section{Learning from Pairwise Comparison}\label{sec:pairwise}

We begin with the problem of learning from pairwise comparisons under the BTL model.

 \subsection{Algorithms: MLE and Pessimistic MLE}
 
We first bound  the estimation error for MLE, the most common algorithm in learning to rank and  RLHF~\cite{liu2009learning, xia2008listwise, cao2007learning, christiano2017deep, ouyang2022training}.  For any query-observation dataset $\{(s^i, a^i_{1},a^i_{2}, y^i)\}_{i=1}^n$, MLE aims at minimizing the negative log likelihood, defined as:
\begin{align}
\hat \theta_{\mathsf{MLE}} & \in \argmin_{\theta\in\Theta_B}    \ell_\mathcal{D}(\theta),\nonumber \\
\ell_\mathcal{D}(\theta) & = -\sum_{i=1}^n \log \Big(\frac{1(y^i=1)\cdot \exp(r_{\theta}(s^i, a_{1}^i))}{\exp(r_{\theta}(s^i, a_{0}^i))+\exp(r_{\theta}(s^i, a_{1}^i))} +  \frac{1(y^i=0)\cdot\exp(r_{\theta}(s^i, a_{0}^i))}{\exp(r_{\theta}(s^i, a^i_{0}))+\exp(r_{\theta}(s^i, a^i_{1}))}\Big)\nonumber \\
& = -\sum_{i=1}^n \log \Big(1(y^i=1)\cdot \mathsf{sigmoid}( \inprod{\theta}{\phi(s^i, a_1^i) - \phi(s^i, a_0^i)})  +1(y^i=0)\cdot \mathsf{sigmoid}( \inprod{\theta}{\phi(s^i, a_0^i) - \phi(s^i, a_1^i)})\Big). \nonumber
\end{align}

When the minimizer is not unique, we take any of the $\hat \theta$ that achieve the minimum. 
Let $\mathcal{D}= \{(s^i, a_{1}^i,a_{2}^i)\}_{i=1}^n$ denote the queried state-action pairs. In  this paper, we study how one can utilize $\mathcal{D}$ to learn a near-optimal reward model and policy. 
We first present a lemma on the estimation error conditioned on the data $\mathcal{D}$. The lemma is a generalization of the upper bound in \citet[Theorem 1]{shah2015estimation} and the analysis follows a similar structure. The main difference is that \citet{shah2015estimation} focus on the tabular case when $\phi(s, a)$ is always a standard basis vector, while in our case $\phi(s, a)$ can be an arbitrary $d$-dimensional vector. This confidence bound guarantee is also similar to the guarantee for dueling bandits and RL in~\citet{faury2020improved, pacchiano2021dueling}, except for that we have better rate in logarithmic factors since union bound is not needed in our case.   

\begin{lemma}\label{lem:mle_estimation}
For any $\lambda>0$,
with probability at least $1-\delta$,
\begin{align*}
\|\hat\theta_{\mathsf{MLE}} - \theta^\star\|_{\Sigma_{\mathcal{D}}+\lambda I}\leq C\cdot \sqrt{ \frac{d+\log(1/\delta)}{\gamma^2 n} +  {\lambda B^2}}.
\end{align*}
Here $\Sigma_\mathcal{D} = \frac{1}{n}\sum_{i=1}^n (\phi(s^i, a_1^i) - \phi(s^i, a_0^i))(\phi(s^i, a_1^i) - \phi(s^i, a_0^i))^\top $, $\strongcon = 1/(2+\exp(-LB)+\exp(LB))$.
\end{lemma}

The proof is deferred to Appendix~\ref{proof:mle_estimation}.  
The optimality of the bound can be seen  via a lower-bound argument akin to that in~\citet[Theorem 1]{shah2015estimation}. 

Now consider the set of parameters 
\begin{align*}
    \Theta(\hat \theta_{\mathsf{MLE}},\lambda) & = \Big\{\theta \in\Theta_B \mid \|\hat \theta_{\mathsf{MLE}} -\theta\|_{\Sigma_{\mathcal{D}}+\lambda I}   \leq C\cdot \sqrt{ \frac{d+\log(\frac{1}{\delta})}{\gamma^2 n} +  {\lambda B^2}}\Big\}.
 \end{align*} 
Lemma~\ref{lem:mle_estimation} shows that with probability at least $1-\delta$, one has $\theta^\star\in\Theta(\hat \theta_{\mathsf{MLE}})$. 
We thus consider the  pessimistic MLE   in Algorithm~\ref{alg:LCB}, which takes the lower confidence bound (LCB)  as the reward estimate.  In the context of LLM, the features of meaningful prompts and responses usually lie on a low-dimensional manifold. The idea of pessimism is to assign larger reward for the responses that lie on the manifold, and penalize the rarely seen responses that do not lie on manifold. 
We have the following guarantee for pessimistic MLE:

\begin{algorithm}[tb]
\caption{Pessimistic MLE}
\label{alg:LCB}
\begin{algorithmic}
\STATE \textbf{Input:} The current estimator $\hat\theta$, the data covariance $\Sigma_{\mathcal{D}}$, the regularization parameter $\lambda$, the bound on the semi-norm $f(n, d, \delta, \lambda)$, a reference vector $v\in\mathbb{R}^d$,  state distribution $q$
\STATE Construct the confidence set 
\begin{align*}
    \Theta(\hat \theta,\lambda) & = \Big\{\theta\in\Theta_B \mid \|\hat \theta -\theta\|_{\Sigma_{\mathcal{D}}+\lambda I}   \leq f(n, d,\delta,\lambda)\Big\}.
 \end{align*}
 \STATE Compute the pessimistic expected value function 
 \begin{align}
\hat J(\pi) & = \min_{\theta\in \Theta(\hat  \theta,\lambda)}
\mathbb{E}_{s\sim q}[\theta^\top (\phi(s, \pi(s))-v)]\nonumber  \\
& = (\mathbb{E}_{s\sim q}[\phi(s,\pi(s))]-v)^\top \hat\theta    - \|(\Sigma_{\mathcal{D}}+\lambda I)^{-\frac{1}{2}}(\mathbb{E}_{s\sim q}[\phi(s,\pi(s))]-v)\|_2 \cdot  f(n, d,\delta,\lambda) \nonumber 
\end{align}
\STATE \textbf{Return:} $
\hat \pi  = \argmax_{\pi} \hat J(\pi).$
\end{algorithmic}
\end{algorithm}

\begin{theorem}\label{thm:LCB_upper}
Let $\hat \pi_{\mathsf{PE}}$ be the output of Algorithm~\ref{alg:LCB} when taking $\hat\theta = \hat\theta_{\mathsf{MLE}}$, $ f(n, d, \delta, \lambda) =  C\cdot \sqrt{ \frac{d+\log(1/\delta)}{\gamma^2 n} +  {\lambda B^2}}$, $q=\rho$. For any $\lambda>0$ and $v\in\mathbb{R}^d$,  
 with probability at least $1-\delta$, 
    \begin{align*}
       \mathsf{SubOpt}(\hat \pi_{\mathsf{PE}}) &\leq C\cdot \sqrt{ \frac{d+\log(1/\delta)}{\gamma^2 n} +  {\lambda B^2}}   \cdot \|(\Sigma_{\mathcal{D}}+\lambda I)^{-1/2}\mathbb{E}_{s\sim \rho}[(\phi(s, \pi^\star(s))-v)]\|_2. 
    \end{align*}
\end{theorem}
The proof is deferred to Appendix~\ref{proof:lcb_pairwise}.
We make several remarks.
\begin{remark}[\textbf{The single concentratability coefficient assumption}]
When $v=0$,
the term $\|(\Sigma_{\mathcal{D}}+\lambda I)^{-1/2}\mathbb{E}_{s\sim \rho}[\phi(s, \pi^\star(s))]\|_2$ is referred to as a ``single concentratability coefficient'', which is assumed to be bounded  in most of   the  literature on offline learning~\citep{rashidinejad2021bridging, li2022pessimism, xie2021policy, zanette2022realizability, zanette2021provable}.  A bounded concentratability coefficient  can be understood as certifying good coverage of the target vector $\mathbb{E}_{s\sim \rho}[\phi(s, \pi^\star(s))]$ from the dataset $\mathcal{D}$ in the feature space. The  performance guarantee also holds  when we replace $\pi^\star$ with any reference policy $\pi$ on both sides. 
\end{remark}

\begin{remark}[\textbf{The choice of $\lambda$}]
 When $\Sigma_{\mathcal{D}}$ is invertible, or when any $\theta\in\Theta_B$ is orthogonal to the nullspace of $\Sigma_{\mathcal{D}}$, the above inequality holds for the case of $\lambda = 0$. In other cases, one may minimize $\lambda$ on the right-hand side, or simply take $\lambda = (d+\log(1/\delta)/(B^2\gamma^2 n))$ to achieve a near-optimal rate up to a constant factor.
\end{remark}

\begin{remark}[\textbf{The choice of $v$}]
Compared to the traditional pessimism principle~\citep{rashidinejad2021bridging, li2022pessimism, xie2021policy, zanette2022realizability, zanette2021provable}, we subtract an extra reference vector $v$ in all the feature vectors $\phi$.  Subtracting a constant vector in feature space will not change the induced policy, but may affect the  concentratability coefficient $\|(\Sigma_{\mathcal{D}}+\lambda I)^{-1/2}(\mathbb{E}_{s\sim \rho}[\phi(s,\pi(s))]-v)\|_2$.  

We briefly describe the reason for introducing $v$ here.  Consider the case where   the differences between features lie in the same subspace, while  the  feature $\phi$ itself does not. As a concrete example, consider a single state $s$ and two actions $a_0, a_1$, we let $\phi(s,a_0)=(1,1)$ and $\phi(s, a_1) = (1,0)$. The data covariance is $(\phi(s^i, a_1^i) - \phi(s^i, a_0^i))(\phi(s^i, a_1^i) - \phi(s^i, a_0^i))^\top =[0,0;0,1]$. Thus $\|(\Sigma_{\mathcal{D}}+\lambda I)^{-1/2}\phi(s,a_0)\|_2$ can be arbitrarily large as $\lambda\rightarrow 0$ when $v=0$. On the other hand, when we take $v=\phi(s, a_1)$, one can verify that $\|(\Sigma_{\mathcal{D}}+\lambda I)^{-1/2}(\phi(s,a_0)-v)\|_2\leq 1$.

The above example illustrates the importance of choosing an appropriate $v$. A good rule of thumb for choosing $v$ is the most common feature vector $\phi$ that appears in the data, so that more features can be covered.  This  also affords additional design latitude for other pessimism algorithms.   
\end{remark}
\begin{remark}[\textbf{Implementation for neural network}]
When $r_\theta$ is a neural network, Algorithm~\ref{alg:LCB} may not be directly implementable.  As an alternative, there has been a  number of heuristic approximations considered, including Conservative Q-Learning~\citep{kumar2020conservative}, Implicit Q-Learning~\citep{kostrikov2021offline} and Adversarially Trained Actor Critic~\citep{cheng2022adversarially}.  Furthermore, one may also introduce pessimism in the policy training procedure. For example, ~\citet{ouyang2022training} add regularization terms in policy training, which enforces that the policy stays close to the original policy, and within the coverage of the pre-trained dataset. 
Our analysis supplies a theoretical rationale for such  regularization terms.
\end{remark}
\begin{remark}[\textbf{Implications for online learning}]
    Although we mainly focus on offline learning, Lemma~\ref{lem:mle_estimation} also gives a straightforward online learning algorithm when combined with an optimism-based algorithm. In particular, a pure exploration-based active learning scheme would seek to compare pairs of actions whose feature difference is poorly covered by the past observations; i.e., find $(s,a_1,a_2)$ such that $\|\phi(s,a_1)-\phi(s, a_2)\|_{(\Sigma_{\mathcal{D}}+\lambda I)^{-1}}$ is maximized. As a corollary of Lemma~\ref{lem:mle_estimation} and exploration results for linear bandits~\citep{abbasi2011improved, soare2014best}, one can derive tight regret bound for online learning. 
\end{remark}

\begin{remark}[\textbf{Special Case: Multi-Armed Bandit}]
For multi-armed bandits we have only a single state, such that the feature $\phi(s, a)$ reduces to $\vec 1_a$, which is a unit vector with $1$ on its $a$-th element. In this case, the data covariance reduces to a Laplacian matrix, defined as 
$
\Sigma_\mathcal{D} = \frac{1}{n}\sum_{i=1}^n (\vec 1_{a_1} - \vec 1_{a_0})(\vec 1_{a_1} - \vec 1_{a_0})^\top.$
This is precisely the problem considered in~\citet{shah2015estimation}. 
The Laplacian matrix is positive semidefinite and always has a zero eigenvalue, corresponding to an all ones eigenvector.  When the graph induced by the Laplacian matrix is connected, any $\theta$ with $\inprod{1}{\theta} =0$ is orthogonal to the nullspace of $\Sigma_{\mathcal{D}}$, thus the theorem holds for the case of $\lambda = 0$.
\end{remark}

\subsection{Failure of MLE and Lower Bounds}
We also show that there exists a simple linear bandit where MLE fails and pessimistic MLE succeeds. Let $\hat\pi_{\mathsf{MLE}}=\argmax_\pi \mathbb{E}[ r_{\hat\theta_{\mathsf{MLE}}}(s,\pi(s))]$ be the greedy policy with respect to the MLE. 
\begin{theorem}\label{thm:fail_mle}
    There exists a linear bandit with four actions and a sampling distribution such that for any $n>1$,\begin{align*}
       \mathbb{E}[ \mathsf{SubOpt}(\hat \pi_{\mathsf{MLE}})]\geq 0.1.
    \end{align*}
On the other hand,  with probability at least $1-\delta$,
    \begin{align*}
 \mathsf{SubOpt}(\hat \pi_{\mathsf{PE}})\leq \frac{C\cdot \log(1/\delta)}{\sqrt{n}}.
    \end{align*}
 Here $C$ is some universal constant.
\end{theorem}
The proof is deferred to Appendix~\ref{proof:fail_mle}. The results show a separation between MLE and pessimistic MLE when the concentratability coefficient is bounded.   The failure of MLE has also been empirically observed in~\citet{gao2022scaling}, which leads to overoptimization with the trained reward model.

We also show that for the  problems with bounded concentratability coefficient, pessimistic MLE is minimax-rate optimal up to a constant factor. 
Consider the family of contextual bandit instances as follows:
\begin{align}
    \mathsf{CB}(\Lambda) & = \{\rho, \{(s_i, a_{i1}, a_{i2})\}_{i=1}^n, \theta^\star \mid \|\Sigma_{\mathcal{D}}^{-1/2}\mathbb{E}_{s\sim\rho}[\phi(s,\pi^\star(s))]\|_2\leq \Lambda  \}.\nonumber 
\end{align}
Here we assume that $\Sigma_{\mathcal{D}}$ is invertible to simplify the presentation of the lower bound.
For any $\mathcal{Q}\in   \mathsf{CB}(\Lambda)$, we let $\mathsf{SubOpt}_\mathcal{Q}(\pi)$ be the sub-optimality under instance $\mathcal{Q}$.  
We have the following lower bound result, the proof  of which is deferred to Appendix~\ref{proof:lcb_lower}.
\begin{theorem}\label{thm:lcb_lower}
For any $d>6, n\geq Cd\Lambda^2, \Lambda\geq 2$, there exists a feature mapping $\phi$ such that the following lower bound holds. 
\begin{align*}
\inf_{\hat \pi} \sup_{\mathcal{Q}\in \mathsf{CB}(\Lambda)} \mathsf{SubOpt}_\mathcal{Q}(\hat\pi)\geq C\Lambda\cdot \sqrt{\frac{d}{n}}.   
\end{align*}
\end{theorem}
Comparing with the upper bound in Theorem~\ref{thm:LCB_upper}, we see that the pessimistic MLE is minimax-optimal up to constant factors for the sub-optimality of induced policy.

\section{Learning from   $\numchoices$-wise comparisons}
\label{Sec:Kwise}

We now consider  learning from $K$-wise comparisons under the PL model. In this case, we design two different estimators based on MLE. One involves directly maximizing the likelihood under the PL model, denoted as $\mathsf{MLE}_K$. The other involves splitting the $K$-wise comparison data with pairwise comparisons and running $\mathsf{MLE}$ for pairwise comparisons. We denote this estimator as $\mathsf{MLE}_2$.

\subsection{Algorithms}
\paragraph{Guarantee for $\mathsf{MLE}_K$.}
Let $\mathcal{D}=\{(s^i,a_{0}^{i},\cdots,a_{K}^{i})\}_{i=1}^n$
be the set of queried states and actions, and the permutation function $\sigma^i$ be the output of the $i$-th query. We can compute its maximum likelihood estimator as
\begin{align*}
\hat \theta_{\mathsf{MLE}_K} & \in \argmin_{\theta\in\Theta_B}    \ell_\mathcal{D}(\theta), \nonumber\\ 
\text{where } \ell_\mathcal{D}(\theta) & = - \frac{1}{n}\sum_{i=1}^n \sum_{j=0}^{K-1}  \log \left(\frac{\exp(\inprod{\theta}{\phi(s^i, a^i_{\sigma_i(j)})} )}{\sum_{k=j}^{\numchoices-1} \exp(\inprod{\theta}{\phi(s^i, a^i_{\sigma_i(k)})})}\right).\nonumber 
\end{align*}

Similar to~\citet{shah2015estimation}, we restrict our attention to $K = \mathcal{O}(1)$ since it is known that  it is difficult for human to compare more than a
small number of items due to a  limited information storage and
processing capacity~\citep{miller1956magical,kiger1984depth, shiffrin1994seven,
  saaty2003magic}. For instance,~\citet{saaty2003magic} recommend
eliciting preferences over {no more than seven options}.  
We have the following result for
$\numchoices$-wise comparisons.
\begin{theorem}
Under the $\numchoices$-wise  PL model, 
 for any $\lambda>0$,
with probability at least $1-\delta$,
\begin{align*}
\|\hat\theta_{\mathsf{MLE}_K} - \theta^\star\|_{\Sigma_{\mathcal{D}}+\lambda I}\leq C\cdot \sqrt{ \frac{K^4 (d+\log(1/\delta))}{\gamma^2 n} +  {\lambda B^2}}.
\end{align*}
 Here  $\Sigma_{\mathcal{D}} =  \frac{2}{K(K-1)\numobs}  (\sum_{i=1}^{\numobs} 
\sum_{j=0}^{\numchoices-1} \sum_{k=j+1}^{K-1} (\phi(s^i,a^i_{j}) - \phi(s^i,a^i_{k}))(\phi(s^i,a^i_{j}) - \phi(s^i,a^i_{k}))^\top)$, and $\strongcon = \exp(-4LB)$. As a consequence, 
let $\hat \pi_{\mathsf{PE}_K}$ be the output of Algorithm~\ref{alg:LCB} when taking $\hat\theta = \hat\theta_{\mathsf{MLE}_K}$, $ f(n, d, \delta, \lambda) =  C\cdot \sqrt{ \frac{K^4 (d+\log(1/\delta))}{\gamma^2 n} +  {\lambda B^2}}$. For any $\lambda>0$ and $v\in\mathbb{R}^d$,
 with probability at least $1-\delta$, 
    \begin{align*}
       \mathsf{SubOpt}(\hat \pi_{\mathsf{PE}_K}) & \leq C\cdot \sqrt{ \frac{K^4 (d+\log(1/\delta))}{\gamma^2 n} +  {\lambda B^2}} \cdot \|(\Sigma_{\mathcal{D}}+\lambda I)^{-1/2}\mathbb{E}_{s\sim \rho}[(\phi(s, \pi^\star(s))-v)]\|_2. 
    \end{align*}
   
\label{thm:kwise}
\end{theorem}
The proof of Theorem~\ref{thm:kwise} is provided in
Appendix~\ref{proof:kwise}. \citet{shah2015estimation} also study the extension from pairwise to $K$-wise comparisons. However, they focus on the setting where only the maximum is selected, where we assume a complete ranking among $K$ items is given. Also, they only provide an expectation bound while we provide a high-probability bound. 

Compared to the pairwise comparison result in Theorem~\ref{thm:LCB_upper}, the covariance matrix $\Sigma_{\mathcal{D}}$ now takes the sum over the feature differences between all pairs of actions among $K$-wise comparisons. As a cost, the right-hand side bound also introduces extra dependence on $K$.   Our bound is likely to be loose in terms of the dependence on $K$. However, since we mainly focus on the case of $K=\mathcal{O}(1)$, such a bound is still near-optimal due to the minimax lower bound for pairwise comparisons. Furthermore, the gap between MLE and pessimistic MLE for sub-optimality still exists since Theorem~\ref{thm:fail_mle} holds as a special case of $K$-wise comparison.


\paragraph{Guarantee for $\mathsf{MLE}_2$}
Besides the standard MLE approach, another option is to   replace the joint distribution of $K$-ranking data with $K(K-1)/2$ pairs of  pairwise comparisons. This can be understood as replacing the true probability in $\mathsf{MLE}_K$ with the   product of marginals:
\begin{align}
\hat \theta_{\mathsf{MLE}_2} & \in \argmin_{\theta\in\Theta_B}    \ell_\mathcal{D}(\theta), \nonumber \\ 
\text{where } \ell_\mathcal{D}(\theta) & = - \frac{1}{n}\sum_{i=1}^n \sum_{j=0}^{K-1} \sum_{k=j+1}^{K-1} \log \left(\frac{\exp(\inprod{\theta}{\phi(s^i, a^i_{\sigma_i(j)})} )}{\exp(\inprod{\theta}{\phi(s^i, a^i_{\sigma_i(j)})} )+\exp(\inprod{\theta}{\phi(s^i, a^i_{\sigma_i(k)})})}\right) \nonumber .
\end{align}

 This estimator is also applied in the current  RLHF  for LLM~\citep[see, e.g.,][]{ouyang2022training}.
We show that it  also leads to a good induced policy, as is shown in the theorem below. 
\begin{theorem}
  Under the $\numchoices$-wise  PL model, 
 for any $\lambda>0$,
with probability at least $1-\delta$,
\begin{align*}
\|\hat \theta_{\mathsf{MLE}_2} - \theta^\star\|_{\Sigma_{\mathcal{D}}+\lambda I}\leq C\cdot \sqrt{ \frac{d+\log(1/\delta)}{\gamma^2 n} +  {\lambda B^2}}.
\end{align*}
Here $\Sigma_{\mathcal{D}} =  \frac{2}{K(K-1)\numobs}  (\sum_{i=1}^{\numobs} 
\sum_{j=0}^{\numchoices-1} \sum_{k=j+1}^{K-1} (\phi(s^i,a^i_{j}) - \phi(s^i,a^i_{k})) (\phi(s^i,a^i_{j}) - \phi(s^i,a^i_{k}))^\top)$, and  $\strongcon = 1/(2+\exp(-2LB)+\exp(2LB))$. As a consequence,
let $\hat \pi_{\mathsf{PE}_2}$ be the output of Algorithm~\ref{alg:LCB} when taking $\hat\theta = \hat\theta_{\mathsf{MLE}_2}$, $ f(n, d, \delta, \lambda) =  C\cdot \sqrt{ \frac{d+\log(1/\delta)}{\gamma^2 n} +  {\lambda B^2}}$, $q=\rho$. For any $\lambda>0$ and $v\in\mathbb{R}^d$,
 with probability at least $1-\delta$, 
    \begin{align*}
       \mathsf{SubOpt}(\hat \pi_{\mathsf{PE}_2}) & \leq C\cdot \sqrt{ \frac{ d+\log(1/\delta)}{\gamma^2 n} +  {\lambda B^2}}  \cdot \|(\Sigma_{\mathcal{D}}+\lambda I)^{-1/2}\mathbb{E}_{s\sim \rho}[(\phi(s, \pi^\star(s))-v)]\|_2. 
    \end{align*} 
\label{thm:kwise_pair}
\end{theorem}
The proof of Theorem~\ref{thm:kwise_pair} is provided in
Appendix~\ref{proof:kwise_pair}. Our theoretical analysis  validates the empirical performance of $\mathsf{MLE}_2$ in~\citet{ouyang2022training}.
Compared to the guarantee for $\mathsf{MLE}_K$, ${\mathsf{MLE}_2}$ seems to has  better nonasymptotic upper bound in terms of the dependence on $K$. However, it is likely that this comes from a loose analysis of $\mathsf{MLE}_K$. The $\mathsf{MLE}_2$ belongs to the family of the M-estimators, whose asymptotic variance is known to be larger than  that of MLE~\cite{godambe1960optimum, lee2008m}. Thus, asymptotically, $\mathsf{MLE}_K$ is more efficient than $\mathsf{MLE}_2$.  We can calculate the asymptotic variance of both estimators  as follows:
\begin{theorem}
    We have
    \begin{align*}
        \sqrt{n}(\hat \theta_{\mathsf{MLE}_K}-\theta^\star) &\rightarrow \mathcal{N}(0, \mathcal{I}(\theta^\star)^{-1}); \\ 
        \sqrt{n}(\hat \theta_{\mathsf{MLE}_2}-\theta^\star) &\rightarrow \mathcal{N}(0, V).
    \end{align*}
    where
    \begin{align*}
        \mathcal{I}(\theta^\star) & = \mathbb{E}_{\theta^\star}\Bigg[\sum_{j=0}^{K-1}\sum_{k=j}^{K-1}\sum_{k'=j}^{K-1}\frac{\exp(\inprod{\theta^\star}{\phi(s^i, a^i_{\sigma_i(k)})+\phi(s^i, a^i_{\sigma_i(k')})})}{(\sum_{k'=j}^{K-1} \exp(\inprod{\theta^\star}{\phi(s^i, a^i_{\sigma_i(k')})}))^2} \\
      & \qquad   \cdot (\phi(s^i, a^i_{\sigma_i(k)}) - \phi(s^i, a^i_{\sigma_i(k')}))(\phi(s^i, a^i_{\sigma_i(k)}) - \phi(s^i, a^i_{\sigma_i(k')}))^\top\Bigg],\\
        V &=\Sigma^{-1}\mathbb{E}_{\theta^\star}\left[GG^\top\right]\Sigma^{-1}, \\
        \Sigma & = \mathbb{E}_{\theta^\star}\left[ \sum_{j=0}^{K-1}\sum_{k=j}^{K-1}\frac{\exp(-\inprod{\theta^\star}{x^i_{\sigma_i(j)\sigma_i(k)})}}{(1+\exp(-\inprod{\theta^\star}{x^i_{\sigma_i(j)\sigma_i(k)})})^2} \cdot \left(\phi(s^i,a^i_{\sigma_i(j)})-\phi(s^i,a^i_{\sigma_i(k)})\right)\left(\phi(s^i,a^i_{\sigma_i(j)})-\phi(s^i,a^i_{\sigma_i(k)})\right)^\top\right], \\
        G & = \sum_{j=0}^{K-1}\sum_{k=j+1}^{K-1}\frac{\exp(-\inprod{\theta^\star}{x^i_{\sigma_i(j)\sigma_i(k)})}}{1+\exp(-\inprod{\theta^\star}{x^i_{\sigma_i(j)\sigma_i(k)})}}\cdot \left(\phi(s^i,a^i_{\sigma_i(j)})-\phi(s^i,a^i_{\sigma_i(k)})\right)
    \end{align*}
\end{theorem}
The proof follows directly the gradient and Hessian computed in Appendix~\ref{proof:kwise} and ~\ref{proof:kwise_pair}, combined with~\citet[Section 5.3]{van2000asymptotic}. We also empirically verify the performances of both estimators in Section~\ref{app:experiments}.

\section{Extension to MDPs}\label{sec:MDP}
Thus far we have considered only contextual bandits. We now extend our results to the MDP setting. Depending on whether the comparison is based on a single action or a whole trajectory, we have two regimes, namely action-based comparison and trajectory-based comparison.  

\subsection{Trajectory-based Comparison}
In trajectory-based comparison, we assume that two trajectories that start from the same initial  state are given, and the comparison is based on the cumulative reward of the two trajectories. 
Concretely,  we first sample the initial state $s_0$ from some fixed distribution $\rho$, and then sample two trajectories  $\tau_0=(a_0, s_1, a_1,\cdots, s_H, a_H)$ and $\tau_1= (a_0', s_1', a_1',\cdots, s_H', a_H')$  from joint distributions $P_l(a_0, s_1, a_1,\cdots, s_H, a_H|s_0)=\prod_i \pi_l(a_i|s_i) P(s_{i+1} | s_{i},a_i)$, where $l\in\{0,1\}$.
For each queried state-trajectory pair, we  observe a sample $y$ from a Bernoulli distribution as follows:
\begin{align}
    \mathbb{P}(y = 1 \mid s, \tau_0, \tau_1) & = \frac{\exp(\sum_{h=0}^H r_{\theta^\star}(s_h, a_{h}))}{\exp(\sum_{h=0}^H r_{\theta^\star}(s_h, a_{h}))+\exp(\sum_{h=0}^H r_{\theta^\star}(s_h', a_{h}')))}. \nonumber  
\end{align}

Given the dataset $\{(s^i, \tau^i_0, \tau^i_1, y^i\}_{i=1}^n$, 
the MLE  is
\begin{align}
\hat \theta_{\mathsf{MLE}} & \in \argmin_{\theta\in\Theta_B}    \ell_\mathcal{D}(\theta), \nonumber \\ 
\text{where } \ell_\mathcal{D}(\theta) & = -\sum_{i=1}^n \log \Big( \frac{1(y^i=1)\cdot\exp(\sum_{h=0}^H r_{\theta}(s^i_h, a_{h}^i))}{\exp(\sum_{h=0}^H r_{\theta}(s^i_h, a_{h}^i))+\exp(\sum_{h=0}^H r_{\theta}(s_h^{i\prime}, a_h^{i\prime}))}  +\frac{1(y^i=0)\cdot \exp(\sum_{h=0}^H r_{\theta}(s_h^{i\prime}, a_h^{i\prime}))}{\exp(\sum_{h=0}^H r_{\theta}(s^i_h, a_{h}^i))+\exp(\sum_{h=0}^H r_{\theta}(s_h^{i\prime}, a_h^{i\prime}))}\Big). \nonumber
\end{align}

Compared to the pairwise comparison in the contextual bandit, the exponent changes from a single reward to the cumulative reward. Similarly, 
 we provide the following guarantee for the estimation error of  MLE: 

\begin{lemma}\label{lem:mle_estimation_rl}
Assume that 
$\|\phi(\cdot, \cdot)\|_\infty\leq L$ for any $s, a$. Then for any $\lambda>0$,
with probability at least $1-\delta$,
\begin{align*}
\|\hat\theta_{\mathsf{MLE}} - \theta^\star\|_{\Sigma_{\mathcal{D}}+\lambda I}\leq C\cdot \sqrt{ \frac{d\log(1/\delta)}{\gamma^2 n} +  {\lambda B^2}}.
\end{align*}
Here $\Sigma_{\mathcal{D}} = \frac{1}{n}\sum_{i=1}^n (\sum_{h=0}^H (\phi(s^i_h, a_{h}^i)-\phi(s_h^{i\prime}, a_{h}^{i\prime})))$   $(\sum_{h=0}^H (\phi(s^i_h, a_{h}^i)-\phi(s_h^{i\prime}, a_{h}^{i\prime})))^\top$, and $\strongcon = 1/(2+\exp(-2HLB)+\exp(2HLB))$. 
\end{lemma}
The proof is deferred to Appendix~\ref{proof:mle_estimation_rl_traj}. Compared to the guarantee for contextual bandits in Lemma~\ref{lem:mle_estimation}, the features in the covariance is now the difference between the cumulative feature in trajectory $\tau$ and the cumulative feature in trajectory $\tau'$. The result reduces to Lemma~\ref{lem:mle_estimation} when $H=1$. 

In order to bound the sub-optimality of the induced policy, one needs to plug-in a pessimistic version of the reward estimate. Note that from the definition of $d^\pi$, one has 
\begin{align*}
    \mathbb{E}_{s\sim\rho}[V^\pi(s)] = \mathbb{E}_{s, a\sim d^{\pi}}[r(s, a)].
\end{align*}
In the case when the transition distribution $P$ is known, one may directly compute $d^\pi$ for any policy $\pi$ and replace the initial distribution $\rho$ in the algorithm for contextual bandit. This gives the following result: 
\begin{theorem}\label{thm:LCB_upper_RL_traj}
Let $\hat \pi_{\mathsf{PE}}$ be the output of Algorithm~\ref{alg:LCB} when taking $\hat\theta = \hat\theta_{\mathsf{MLE}}$, $ f(n, d, \delta, \lambda) =  C\cdot \sqrt{ \frac{ d+\log(1/\delta)}{\gamma^2 n} +  {\lambda B^2}}$, $q=d^\pi$. For any $\lambda>0$ and $v\in\mathbb{R}^d$,
 with probability at least $1-\delta$, 
    \begin{align*}
       \mathsf{SubOpt}(\hat \pi_{\mathsf{PE}}) & \leq C\cdot \sqrt{ \frac{d+\log(1/\delta)}{\gamma^2 n} +  {\lambda B^2}}\\ 
       \quad & \cdot \|(\Sigma_{\mathcal{D}}+\lambda I)^{-1/2}\mathbb{E}_{s\sim d^{\pi^\star}}[(\phi(s, \pi^\star(s))-v)]\|_2. 
    \end{align*}
\end{theorem}

The proof is deferred to Appendix~\ref{proof:LCB_upper_rl_traj}.  The result can be  generalized to the case of $K$-wise comparisons following the same argument in Section~\ref{Sec:Kwise}.

\subsection{Action-based Comparison}\label{app:action-based}

In action-based comparison, we assume that two actions are sampled for each state, and the comparison is based on the expected cumulative return starting from such state-action pair. 

Concretely, 
assume that the optimal $Q$-function is parameterized as $Q_\theta^\star(s, a) = \theta^\top \phi(s, a)$ for some given $\phi(s, a)$. Let $\theta^\star$ be the true parameter. During the training, we first sample the state $s$ from some fixed distribution $\rho$, and then sample a pair of actions $a_0, a_1$  from a joint distribution $P(a_0, a_1|s)$. For each queried state-actions pair $(s, a_0, a_1)$, we  observe a sample $y$ from a Bernoulli distribution with parameter $\frac{\exp(Q_{\theta^\star}(s, a_{1}))}{\exp(Q_{\theta^\star}(s, a_{0}))+\exp(Q_{\theta^\star}(s, a_{1}))}$, i.e.
\begin{align*}
    \mathbb{P}(y = 1 \mid s, a_0, a_1) = \frac{\exp(Q_{\theta^\star}(s, a_{1}))}{\exp(Q_{\theta^\star}(s, a_{0}))+\exp(Q_{\theta^\star}(s, a_{1}))} \quad \text{and} \quad  \mathbb{P}(y = 0 \mid s, a_0, a_1) = \frac{\exp(Q_{\theta^\star}(s, a_{0}))}{\exp(Q_{\theta^\star}(s, a_{0}))+\exp(Q_{\theta^\star}(s, a_{1}))}. 
\end{align*}
In this case, one may use the same MLE to estimate $\theta^\star$, which results in an estimator $\hat Q$ for the $Q^\star$-function.  
The following lemma follows exactly the same analysis as Lemma~\ref{lem:mle_estimation}:

\begin{lemma} 
Under the BTL model for action-based RLHF, 
 for any $\lambda>0$,
with probability at least $1-\delta$,
\begin{align*}
\|\hat\theta_{\mathsf{MLE}} - \theta^\star\|_{\Sigma_{\mathcal{D}}+\lambda I}\leq C\cdot \sqrt{ \frac{ d+\log(1/\delta)}{\gamma^2 n} +  {\lambda B^2}}.
\end{align*}
Here $\strongcon = 1/(2+\exp(-LB)+\exp(LB))$.
$\Sigma_\mathcal{D} = \frac{1}{n}\sum_{i=1}^n (\phi(s^i, a_1^i) - \phi(s^i, a_0^i))(\phi(s^i, a_1^i) - \phi(s^i, a_0^i))^\top $. 
\end{lemma}
When $\Sigma_{\mathcal{D}}$ is invertible and covers all the directions well, this will lead to a valid confidence bound for $Q^\star$, which  implies a good performance of the induced greedy policy without pessimism. 
However, when $\Sigma_{\mathcal{D}}$ does not provide good coverage, introducing pessimism in this case can be hard. The reason is that one needs to construct lower confidence bound for $Q^\pi$ for any $\pi$. However, given such confidence bound of $\hat\theta_{\mathsf{MLE}}$, one can only construct confidence bound for $Q^\star$. 

\section{Connection with Inverse Reinforcement Learning}\label{sec:irl}
In Inverse Reinforcement Learning (IRL), BTL and PL model are also popular model of human behavior. However, in IRL it is assumed that we only observe the human behavior, which is sampled from the distribution under PL model.   Thus no comparison is queried. Depending on the comparison is action-based on trajectory-based, one has max-entropy IRL or action-based IRL, discussed in details below.

\subsection{Trajectory-based IRL}
In max-entropy IRL~\cite{ziebart2008maximum}, it is also assumed that the human selection of trajectory follows a PL model. 
A common assumption in IRL or IL is that the observed trajectory collected by human behavior is likely to be the optimal policy.  Assumee that the transitions  are deterministic. For any trajectory $\tau=(s_0, a_0,\cdots, s_H,a_H)$, it is assumed that the expert chooses trajectory $\tau$ under  the following model:
\begin{align*}
    \mathbb{P}(\tau) = \frac{\exp(\sum_{h=0}^H \inprod{\theta^\star}{\phi(s_h, a_h)})}{\sum_{\tau'\in\mathcal{T}(s_0)} \exp(\sum_{h=0}^H \inprod{\theta^\star}{\phi(s_h', a_h')})}.
\end{align*}

Here the set $\mathcal{T}(s_0)$   denotes the set for  all possible trajectories that start from $s_0$. Each trajectory is represented by $\tau' = \{(s_h', a_h')\}_{h=1}^H$.  Assume that we are given a set of trajectories $\{s_h^i, a_h^i\}_{i\in[n],h\in[H]}$ that are sampled from the distribution $  \mathbb{P}(\tau) $. When the denominator can be computed exactly, the algorithm of max entropy IRL also reduces to the MLE, which can be written as
\begin{align*}
\hat \theta_{\mathsf{MLE}} & \in \argmin_{\theta\in\Theta_B}    \ell_\mathcal{D}(\theta), \\ 
\text{where } \ell_\mathcal{D}(\theta) & = - \frac{1}{n}\sum_{i=1}^n  \log \left(\frac{\exp(\sum_{h=0}^H \inprod{\theta}{\phi(s_h^i, a_h^i)})}{\sum_{\tau'\in\mathcal{T}(s_0^i)} \exp(\sum_{h=0}^H \inprod{\theta}{\phi(s_h', a_h')})}\right).
\end{align*}
Although the enumeration of all trajectories $\mathcal{T}(s_0^i)$ is not possible due to exponential growth of the possible trajectories with respect to horizon $H$, ~\citet{ziebart2008maximum} provides an alternative way of computing the gradient via calculating the expected state frequency. This enables the  efficient implementation of MLE. 
One can show the performance guarantee for max entropy IRL as follows:

\begin{lemma} \label{lem:mle_estimation_IRL}
Under the PL model, 
 for any $\lambda>0$,
with probability at least $1-\delta$,
\begin{align*}
\|\hat\theta_{\mathsf{MLE}} - \theta^\star\|_{\Sigma_{\mathcal{D}}+\lambda I}\leq C\cdot \sqrt{ \frac{\sup_s|\mathcal{T}(s)|^2\cdot(d+\log(1/\delta))}{\gamma^2 n} +  {\lambda B^2}}.
\end{align*}
Here  $\Sigma_{\mathcal{D}} =  \frac{1}{\numobs\sup_s|\mathcal{T}(s)|^2}  \sum_{i=1}^{\numobs} \sum_{\{(s_h,a_h)\}\in\mathcal{T}(s_0^i)} \sum_{\{(s^{'}_h,a^{'}_h)\}\in\mathcal{T}(s_0^i)} 
(\sum_{h=0}^H (\phi(s_h, a_{h})-\phi(s_h^{\prime}, a_{h}^{\prime})))(\sum_{h=0}^H (\phi(s_h, a_{h})-\phi(s_h^{\prime}, a_{h}^{\prime})))^\top$, and $\strongcon = \exp(-4LB)/2$.
\end{lemma}

Given such guarantee for MLE, we also show that IRL, when combined with pessimism principle, will lead to a good policy. 

\begin{theorem}\label{thm:LCB_upper_IRL}
Let $\hat \pi_{\mathsf{PE}}$ be the output of Algorithm~\ref{alg:LCB} when taking $\hat\theta = \hat\theta_{\mathsf{MLE}}$, $ f(n, d, \delta, \lambda) =  C\cdot \sqrt{ \frac{\sup_s|\mathcal{T}(s)|(d+\log(1/\delta))}{\gamma^2 n} +  {\lambda B^2}}$, $q=d^\pi$. For any $\lambda>0$ and $v\in\mathbb{R}^d$,
 with probability at least $1-\delta$, 
    \begin{align*}
       \mathsf{SubOpt}(\hat \pi_{\mathsf{PE}}) & \leq C\cdot \sqrt{ \frac{\sup_s|\mathcal{T}(s)|^2 (d+\log(1/\delta))}{\gamma^2 n} +  {\lambda B^2}}\\ 
       & \quad \cdot \|(\Sigma_{\mathcal{D}}+\lambda I)^{-1/2}\mathbb{E}_{s\sim \rho}[(\phi(s, \pi^\star(s))-v)]\|_2. 
    \end{align*}
\end{theorem}

The proof of  Lemma~\ref{lem:mle_estimation_IRL} and Theorem~\ref{thm:LCB_upper_IRL} is provided in
Appendix~\ref{proof:LCB_upper_IRL}. For IRL we have the dependence of $\sup_s|\mathcal{T}(s)|$ in our bound, which can be much larger than $d$.  Similar to the case of $K$-wise comparison, one may also split the one observation into $\sup_s|\mathcal{T}(s)|$ pairwise comparisons, which can help improve the dependence on   $\sup_s|\mathcal{T}(s)|$ in the current analysis.

\subsection{Action-based IRL}
Similar to action-based RLHF, 
action-based IRL also models human choice based on $Q^\star$ instead of cumulative reward~\cite{ramachandran2007bayesian, neu2009training, florence2022implicit}. 
Concretely, 
the human behavior is assumed to be based on the $Q$ function $Q^\star(s, a)= \inprod{\theta^\star}{\phi(s, a)}$, i.e. 
\begin{align*}
    \pi^\star(a|s) =  \frac{\exp(\inprod{\theta^\star}{\phi(s, a)})}{\sum_{a'\in\mathcal{A}} \exp(  \inprod{\theta^\star}{\phi(s, a')})}.
\end{align*}
Here the denominator takes all possible actions. Unlike RLHF where a pair of actions are observed, in IRL or IL, only a single human behavior is observed in each round and there is no comparison, i.e. the observed actions $a$ are sampled from $\pi^\star(a\mid s)$.  Given such observation, one can still run MLE and gives similar performance guarantee. In particular, the MLE is given by 
\begin{align*}
\hat \theta_{\mathsf{MLE}} & \in \argmin_{\theta\in\Theta_B}    \ell_\mathcal{D}(\theta), \\ 
\text{where } \ell_\mathcal{D}(\theta) & = - \frac{1}{n}\sum_{i=1}^n  \log \left(\frac{\exp(\inprod{\theta}{\phi(s^i, a^i)})}{\sum_{a'\in\mathcal{A}} \exp(  \inprod{\theta}{\phi(s^i, a')})}\right).
\end{align*} 
The following lemma follows a similar  analysis as Lemma~\ref{lem:mle_estimation} and Lemma~\ref{lem:mle_estimation_IRL}:

\begin{lemma} 
Under the PL model for action-based IRL, 
 for any $\lambda>0$,
with probability at least $1-\delta$,
\begin{align*}
\|\hat\theta_{\mathsf{MLE}} - \theta^\star\|_{\Sigma_{\mathcal{D}}+\lambda I}\leq C\cdot \sqrt{ \frac{ |\mathcal{A}|^2(d+\log(1/\delta))}{\gamma^2 n} +  {\lambda B^2}}.
\end{align*}
Here $\Sigma_{\mathcal{D}} =  \frac{1}{\numobs |\mathcal{A}|^2}  \sum_{i=1}^{\numobs} \sum_{a\in\mathcal{A}} \sum_{a'\in\mathcal{A}} 
(\phi(s^i, a)-\phi(s^i, a^{\prime}))(\phi(s^i, a)-\phi(s^{i}, a^{\prime})))^\top$, and $\strongcon = \exp(-4LB)/2$.
\end{lemma}
Similar to the case of action-based RLHF, it remains an interesting open problem how one can introduce provable lower confidence bound algorithm for policy learning. 

\section{Experiments}\label{app:experiments}
There has been a large amount of empirical work that demonstrates the success of MLE and pessimistic MLE in RLHF for game playing~\citep{knox2008tamer, macglashan2017interactive, christiano2017deep, warnell2018deep}, robotics~\citep{brown2019extrapolating, shin2023benchmarks} and  language models~\citep{ziegler2019fine, stiennon2020learning, wu2021recursively, nakano2021webgpt,ouyang2022training, menick2022teaching, glaese2022improving, gao2022scaling, bai2022training, ganguli2022red, ramamurthy2022reinforcement}.  
Notably, the concurrent work \citet{shin2023benchmarks} proposes Offline Preference-Based Reward Learning (OPRL), which   trains pessimistic policy from the learned reward and shows empirically the superior performance of pessimistic based method (which can be viewed as an approximation of pessimistic MLE). 

\begin{figure}[!htbp]
     \centering
     \begin{subfigure}[b]{0.45\textwidth}
         \centering
\includegraphics[width=\textwidth]{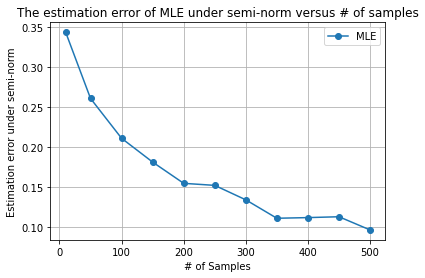}
     \end{subfigure}
     \begin{subfigure}[b]{0.45\textwidth}
         \centering
         \includegraphics[width=\textwidth]{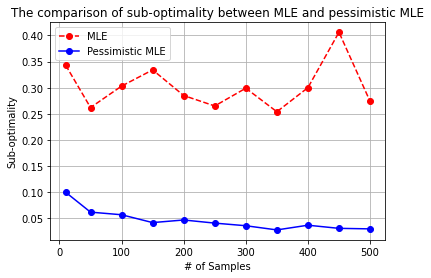}
     \end{subfigure}
        \caption{Left: the convergence of $\mathsf{MLE}$ under the semi-norm $\|\cdot\|_\Sigma$; Right: the comparison between $\mathsf{MLE}$ and pessimistic $\mathsf{MLE}$ under sub-optimality metric.}
        \label{fig:sim}
        \vspace{-5pt}
\end{figure}
In this section, we provide experiments for the contextual bandit case. In particular, we conduct both MLE and pessimistic MLE on the example constructed in Appendix~\ref{proof:fail_mle}. The results are included in Fig.~\ref{fig:sim}. We range the number of samples $n$ from $10$ to $500$. Each sample size is repeated $100$ times. The result verifies our theoretical analysis: MLE converges under the semi-norm but fails to give good policy. On the other hand, pessimistic MLE gives vanishing rate when considering the sub-optimality of the induced policy. Note that in the left figure we do not include pessimistic MLE, since both MLE and pessimistic MLE rely on the same parameter $\hat\theta_{\mathsf{MLE}}$, and they only defer in how the induced policy is trained.

\begin{figure}[!htbp]
\centering
     \begin{subfigure}[b]{0.45\textwidth}
         \centering
\includegraphics[width=\textwidth]{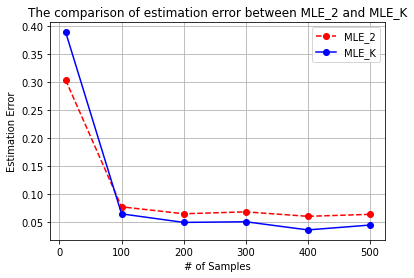}
     \end{subfigure}
     \begin{subfigure}[b]{0.45\textwidth}
         \centering
         \includegraphics[width=\textwidth]{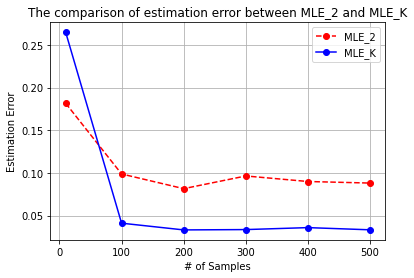}
     \end{subfigure}
    \caption{The comparison of estimation error between $\mathsf{MLE}_2$ and $\mathsf{MLE}_K$, with $K=4$ in the left and $K=9$ in the right.}
    \label{fig:Kwise}
           \vspace{-5pt} 
\end{figure}

On the other hand, we compare the performance of $\mathsf{MLE}_2$ and $\mathsf{MLE}_K$ when learning from $K$-wise comparisons. We take $K=4$ and $K=9$, and range samples from $10$  to $500$. We randomly generate $\phi$ and $\theta^\star$ as independent samples from $3$-dimensional Gaussian distribution. The result is shown in Figure~\ref{fig:Kwise}. One can see that as $n$ grows larger, both estimators converge, while $\mathsf{MLE}_K$ has smaller estimation error than $\mathsf{MLE}_2$. The gap grows larger when $K$ becomes larger. This is consistent with our theoretical prediction in Section~\ref{Sec:Kwise}: since $\mathsf{MLE}_K$ is the true MLE and $\mathsf{MLE}_2$ belongs to the family of M-estimators, asymptotically $\mathsf{MLE}_K$ shall be more efficient than $\mathsf{MLE}_2$.

\section{Conclusion}
We have provided a theoretical analysis of the sample complexity of RLHF. Our main results involve two insights: (i)  pessimism is important to guarantee a good policy; (ii) in $K$-wise comparison, both $\mathsf{MLE}_K$ and   $\mathsf{MLE}_2$ converge. Moreover,  $\mathsf{MLE}_K$ is asymptotically more efficient.

While we have made progress in understanding the reward learning aspect of RLHF, there are many additional questions that remain to be answered. 
\begin{enumerate}
    \item We assumed that the policy trained is greedy with respect to the learned reward. However, in practice the reward is mostly used to fine-tune the pre-trained policy. This requires a more extensive theory that considers the whole procedure of pre-training the policy, learning a reward model and then fine-tuning the policy with policy gradient or PPO. 
 
\item Although we focused on the BTL and PL models, there have been a  number of other models considered for the modeling of human behavior, including the Thurstone model and cardinal models. It would be interesting to extend our analysis to cover these additional models and begin to provide a general characterization of behavioral models for RLHF.  
\item Our constructed confidence bound is based on a fixed feature $\phi$. In the practical fine-tuning scenario, $\phi$ is not fixed but may change slowly. It is interesting to see how the constructed confidence bound helps in the practical fine-tuning scenario for online (active) learning or offline learning, and how one can design valid confidence bound for slowly changing $\phi$.
\end{enumerate}

\section*{Acknowledgement}
The authors would like to thank the reviewers for the valuable suggestions. The authors would like to thank Kihui Lee for pointing out the mistakes in the proof of Theorem 3.10 in an early version of the paper. Banghua Zhu and Jiantao Jiao were partially supported by NSF Grants IIS-1901252 and CCF-1909499. Michael I. Jordan was partially supported by NSF Grants IIS-1901252.
\newpage
\bibliography{ref}

\newpage
\appendix
\onecolumn

\input{appendix.tex}
\end{document}

%% file: macro.tex
\newcommand{\numobs}{\ensuremath{n}}

\def\E{{\mathbb{E}}}

%% file: appendix.tex
\section{Analysis for nonlinear $r_\theta$}\label{app:nonlinear}
Consider the case of pairwise comparison when   $r_\theta$ is not linear, the MLE can be written as
\begin{align*}
\hat \theta_{\mathsf{MLE}} & \in \argmin_{\theta\in\Theta_B}    \ell_\mathcal{D}(\theta), \\ 
\text{where } \ell_\mathcal{D}(\theta) & = -\sum_{i=1}^n \log \left(1(y^i=1)\cdot \frac{\exp(r_{\theta}(s^i, a_1^i))}{\exp(r_{\theta}(s^i, a_0^i))+\exp(r_{\theta}(s^i, a_1^i))}+1(y^i=0)\cdot \frac{\exp(r_{\theta}(s^i, a_0^i))}{\exp(r_{\theta}(s^i, a_0^i))+\exp(r_{\theta}(s^i, a_1^i))}\right).
\end{align*}
Here we provide a guarantee for the case when $r_\theta$ is nonlinear and non-convex. We first make the following boundedness and smoothness assumption on $r_\theta$:
\begin{assumption}\label{ass:nonconvex}
Assume that for any  $\theta\in\Theta_B, s\in\mathcal{S}, a_0\in\mathcal{A},a_1\in\mathcal{A}$ with $a_0\neq a_1$, we have,
\begin{align*}
    |r_\theta(s, a)|& \leq \alpha_0, \quad (\text{Bounded value})\\
     \| \nabla r_\theta(s, a)\|_2& \leq \alpha_1, \quad (\text{Bounded gradient}) \\
    \| \nabla^2 r_\theta(s, a)\|_2& \leq \alpha_2. \quad (\text{Bounded Hessian / Lipschitz gradient}) 
\end{align*}
\end{assumption}
One can verify that our linear reward satisfies the above assumption with $\alpha_0=LB, \alpha_1=L, \alpha_2=0$.  
Under this assumption, we have
\begin{theorem}\label{thm:nonconvex}
  For any $\lambda>0$, with probability at least $1-\delta$,
    \begin{align*}
\|\hat\theta_{\mathsf{MLE}} - \theta^\star\|_{\Sigma_{\mathcal{D}}+\lambda I}\leq C\cdot \sqrt{ \frac{d+\log(1/\delta)}{\gamma^2 n} +  {(\lambda+\alpha_2/\gamma+\alpha_1\alpha_2 B)   B^2}}.
\end{align*}
Here $\strongcon = \frac{1}{2+\exp(-2\alpha_0)+\exp(2\alpha_0)}$, $\Sigma_{\mathcal{D}} = \frac{1}{n}\sum_{i=1}^n \nabla (r_{\theta^\star}(s^i,a_1^i) - r_{\theta^\star}(s^i,a_0^i))\nabla (r_{\theta^\star}(s^i,a_1^i) - r_{\theta^\star}(s^i,a_0^i))^\top$. 
\end{theorem}
The proof is deferred to Appendix~\ref{proof:nonconvex}. Our result recovers Lemma~\ref{lem:mle_estimation} when $\alpha_2=0$ and reveals how the gradient of $r$ plays a role in the bound for estimation error.  However, the dependence on $\alpha_2$ will not vanish as $n\rightarrow \infty$.  It remains open how to get vanishing rate for nonlinear reward functions when $\alpha_2>0$. Similar argument can also be applied to the case of $K$-wise comparison and MDP. And we can similarly design pessimistic MLE based on the confidence bound on $\hat \theta_{\mathsf{MLE}}$.

On the other hand, we can show that the true parameter $\theta^\star$ is a global minimum of the population negative log likelihood even when $r_\theta$ is nonlinear and  we use $\mathsf{MLE}_2$ for $K$-wise comparison. Recall that the $\mathsf{MLE}_2$ splits $K$-wise comparisons into pairwise comparisons, and is given by 
\begin{align*}
\hat \theta_{\mathsf{MLE}_2} & \in \argmin_{\theta\in\Theta_B}    \ell_\mathcal{D}(\theta), \nonumber \\ 
\text{where } \ell_\mathcal{D}(\theta) & = - \frac{1}{n}\sum_{i=1}^n \sum_{j=0}^{K-1} \sum_{k=j+1}^{K-1}  \log \left(\frac{\exp(r_{\theta}{(s^i, a^i_{\sigma_i(j)})} )}{\exp(r_{\theta}{(s^i, a^i_{\sigma_i(j)})} )+\exp(r_{\theta}{(s^i, a^i_{\sigma_i(k)})})}\right) \nonumber .
\end{align*}
When there is infinite number of data, the loss become
\begin{align*}
    \mathbb{E}[\loss(\theta)] & = - \sum_{s} \rho(s) \sum_{a_0, a_1\in\mathcal{A}} \rho(a_0, a_1\mid s) \cdot \Big(  \frac{\exp(r_{\theta^\star}{(s, a_{0})} )}{\exp(r_{\theta^\star}{(s, a_0)} )+\exp(r_{\theta^\star}{(s, a_1)})}\log \Big(\frac{\exp(r_{\theta}{(s, a_{0})} )}{\exp(r_{\theta}{(s, a_0)} )+\exp(r_{\theta}{(s, a_1)})}\Big)   \\
    & \quad +\frac{\exp(r_{\theta^\star}{(s, a_{1})} )}{\exp(r_{\theta^\star}{(s, a_0)} )+\exp(r_{\theta^\star}{(s, a_1)})}\log \Big(\frac{\exp(r_{\theta}{(s, a_{1})} )}{\exp(r_{\theta}{(s, a_0)} )+\exp(r_{\theta}{(s, a_1)})}\Big)    \Big). 
\end{align*}
Here $\rho(a_0, a_1 \mid s)$ is the probability that actions $a_0, a_1$ are included in the $K$-comparison when the state is $s$. Now we show  
\begin{align}\label{eqn:theta_optimal}
    \theta^\star\in \argmin_{\theta} \mathbb{E}[\loss(\theta)].
\end{align}
To see this, note that we have
\begin{align*}
    \mathbb{E}[\loss(\theta)] & = - \sum_{s} \rho(s) \sum_{a_0, a_1\in\mathcal{A}} \rho(a_0, a_1\mid s) \cdot \Big(  \frac{\exp(r_{\theta^\star}{(s, a_{0})} )}{\exp(r_{\theta^\star}{(s, a_0)} )+\exp(r_{\theta^\star}{(s, a_1)})}\log \Big(\frac{\exp(r_{\theta}{(s, a_{0})} )}{\exp(r_{\theta}{(s, a_0)} )+\exp(r_{\theta}{(s, a_1)})}\Big)   \\
    & \quad +\frac{\exp(r_{\theta^\star}{(s, a_{1})} )}{\exp(r_{\theta^\star}{(s, a_0)} )+\exp(r_{\theta^\star}{(s, a_1)})}\log \Big(\frac{\exp(r_{\theta}{(s, a_{1})} )}{\exp(r_{\theta}{(s, a_0)} )+\exp(r_{\theta}{(s, a_1)})}\Big)    \Big) \\ 
    & = \sum_{s} \rho(s) \sum_{a_0, a_1\in\mathcal{A}} p(a_0, a_1\mid s)\cdot (H(p_{\theta^\star}(s,a_0,a_1)) + \mathsf{KL}(p_{\theta^\star}(s,a_0,a_1)\| p_{\theta}(s,a_0,a_1))).
\end{align*}
Here $H(p) = p\log(1/p)+(1-p)\log(1/(1-p))$ is the entropy of a Bernoulli distribution with parameter $p$. And  $p_\theta(s, a_0, a_1) = \frac{\exp(r_{\theta}{(s, a_{1})} )}{\exp(r_{\theta}{(s, a_0)} )+\exp(r_{\theta}{(s, a_1)})}$. Now note that $\mathsf{KL}$ is lower bounded by $0$, with equality when $ \theta = \theta^\star$. This proves Equation~\eqref{eqn:theta_optimal}.

\section{Remaining Proofs}
\subsection{Proof of Lemma~\ref{lem:mle_estimation}}\label{proof:mle_estimation}

Recall that the MLE is given by 

\begin{resizealign}
\hat \theta_{\mathsf{MLE}} & \in \argmin_{\theta\in\Theta_B}    \ell_\mathcal{D}(\theta), \nonumber \\ 
\text{where } \ell_\mathcal{D}(\theta) & = -\sum_{i=1}^n \log \left(1(y^i=1)\cdot \frac{\exp(r_{\theta}(s^i, a_1^i))}{\exp(r_{\theta}(s^i, a_0^i))+\exp(r_{\theta}(s^i, a_1^i))}+1(y^i=0)\cdot \frac{\exp(r_{\theta}(s^i, a_0^i))}{\exp(r_{\theta}(s^i, a_0^i))+\exp(r_{\theta}(s^i, a_1^i))}\right) \nonumber \\ 
&= -\sum_{i=1}^n \log \left(1(y^i=1)\cdot \frac{1}{1+\exp(r_{\theta}(s^i, a_0^i) - r_\theta(s^i, a_1^i))}+1(y^i=0)\cdot \left(1-\frac{1}{1+\exp(r_{\theta}(s^i, a_0^i) - r_\theta(s^i, a_1^i))}\right)\right) \nonumber \\
& = -\sum_{i=1}^n \log \left(1(y^i=1)\cdot \frac{1}{1+\exp(\theta^\top(\phi(s^i, a_0^i) - \phi(s^i, a_1^i)))}+1(y^i=0)\cdot \left(1-\frac{1}{1+\exp(\theta^\top(\phi(s^i, a_0^i) - \phi(s^i, a_1^i)))}\right)\right)\nonumber 
\end{resizealign}

To simplify the notation, we let $x_i = \phi(s^i, a_1^i) - \phi(s^i, a_0^i)$.
Our goal is to bound the estimation error 
of the MLE in the squared semi-norm
$\|v\|_{\Sigma_\mathcal{D}+\lambda I}^2 = {v^\top (\Sigma_\mathcal{D}+\lambda I) v}$.

\paragraph{Strong convexity of $\ell$.}

We first show that $\loss_\mathcal{D}$ is strongly convex at $\theta^\star$ with respect to the semi-norm $\|\cdot\|_{\Sigma_{\mathcal{D}}}$,
meaning that there is some constant $\strongcon > 0$ such that
\begin{align}
\loss_\mathcal{D}(\theta^\star + \Delta) - \loss_\mathcal{D}(\theta^\star) - \inprod{\nabla
  \loss_\mathcal{D}(\theta^\star)}{\Delta} & \geq \strongcon \|\Delta\|_{\Sigma_{\mathcal{D}}}^2
\end{align}
for all perturbations $\Delta \in \real^\numitems$ such that $\theta^\star
+ \Delta\in\Theta_B$.

One can directly calculate the Hessian of $\ell$ as
\begin{align*}
\nabla^2 \loss_\mathcal{D}(\theta) & = \frac{1}{\numobs}
\sum_{i=1}^{\numobs} \left(1(y^i=1)\cdot\frac{\exp(-\inprod{\theta}{\diff_i})}{(\exp(-\inprod{\theta}{\diff_i})+1)^2} +1(y^i=0)\cdot\frac{\exp(\inprod{\theta}{\diff_i})}{(\exp(\inprod{\theta}{\diff_i})+1)^2} \right)\cdot \diff_i \diff_i^\top \\
& = \frac{1}{\numobs}
\sum_{i=1}^{\numobs} \frac{\exp(-\inprod{\theta}{\diff_i})}{(\exp(-\inprod{\theta}{\diff_i})+1)^2} \cdot \diff_i \diff_i^\top
\end{align*}
Observe that  $\inprod{\theta}{\diff_i}\in[-2LB,2LB]$, which gives that  
\begin{align*}
    \frac{\exp(-\inprod{\theta}{\diff_i})}{(\exp(-\inprod{\theta}{\diff_i})+1)^2} \geq \frac{1}{2+\exp(-2LB)+\exp(2LB)}.
\end{align*}
Putting together the pieces, we conclude
that
\begin{align*}
 v^\top \nabla^2 \loss_\mathcal{D} (\theta) v \geq \frac{\strongcon}{\numobs
   } \|X v\|_2^2 \qquad \mbox{for all $v$},
\end{align*}
where $\strongcon = 1/(2+\exp(-2LB)+\exp(2LB))$, $\Xmat \in \real^{\numobs \times \numitems}$ has the
differencing vector $x_i \in \real^\numitems$ as its $i^{th}$ row.
Thus, if we introduce the error vector $\Delta \coloneqq \hat\theta_{\mathsf{MLE}} -
\theta^\star$, then we may conclude that
\begin{align*}
\loss_\mathcal{D}(\theta^\star + \Delta) - \loss_\mathcal{D}( \theta^\star ) - \inprod{\nabla \loss_\mathcal{D}
  (\theta^\star)}{ \Delta} & \geq \frac{\strongcon}{\numobs }
\|X \Delta\|_2^2 \; = \; {\strongcon}{}
\|{\Delta}\|_{\Sigma_{\mathcal{D}}}^2,
\end{align*}
showing that $\loss_\mathcal{D}$ is strongly convex around $\theta^\star$ with
parameter ${\strongcon}$.

\paragraph{Bounding the estimation error.} Now we aim at bounding the estimation error $\LAPNORM{\hat\theta_{\mathsf{MLE}} - \theta^\star}$. 
Since $\hat\theta_{\mathsf{MLE}}$ is optimal for $\loss_\mathcal{D}$, we have $\loss_\mathcal{D}(\hat\theta_{\mathsf{MLE}}) \leq
\loss_\mathcal{D}(\theta^\star)$.  (When $\hat\theta_{\mathsf{MLE}}$ is approximately optimal, i.e. $\loss_\mathcal{D}(\hat\theta_{\mathsf{MLE}}) \leq \min_\theta
\loss_\mathcal{D}(\theta)+\epsilon$, the same argument also holds up to an extra additive term $\epsilon$.) Defining the error vector $\Delta = \hat\theta_{\mathsf{MLE}} -
\theta^\star$, adding and subtracting the quantity $\inprod{\nabla
  \loss_\mathcal{D}(\theta^\star)}{\Delta}$ yields the bound
\begin{align*}
\loss_\mathcal{D}(\theta^\star + \Delta) - \loss_\mathcal{D}(\theta^\star) - \inprod{\nabla
  \loss_\mathcal{D}(\theta^\star)}{\Delta} & \leq - \inprod{\nabla
  \loss_\mathcal{D}(\theta^\star)}{\Delta}.
\end{align*}
By the $\strongcon$-convexity condition, the left-hand side is lower
bounded by $\strongcon \LAPNORM{\Delta}^2$.  As for the right-hand
side, note that $|\inprod{\nabla
  \loss_\mathcal{D}(\theta^\star)}{\Delta}| \leq \|{\nabla \loss_\mathcal{D}(\theta^\star)}\|_{(\Sigma_{\mathcal{D}}+\lambda I)^{-1}} \;
\|{\Delta}\|_{\Sigma_{\mathcal{D}}+\lambda I}$ for any $\lambda>0$. Altogether we have
\begin{align*}
    \strongcon \LAPNORM{\Delta}^2\leq \|{\nabla \loss_\mathcal{D}(\theta^\star)}\|_{(\Sigma_{\mathcal{D}}+\lambda I)^{-1}} \;
\|{\Delta}\|_{\Sigma_{\mathcal{D}}+\lambda I}. 
\end{align*}

Now we further bound the term $\|{\nabla \loss_\mathcal{D}(\theta^\star)}\|_{(\Sigma_{\mathcal{D}}+\lambda I)^{-1}}$. 
Observe that the gradient takes the form
\begin{align*}
\nabla \loss_\mathcal{D} (\theta^\star) & = \frac{-1}{\numobs }
\sum_{i=1}^{\numobs} \left[ \Ind[\obs^i = 1] \frac{
    \exp(-\inprod{\theta^\star}{\diff_i})}{1+\exp(-\inprod{\theta^\star}{\diff_i}))} - \Ind[\obs^i=0] \frac{
    1}{1+\exp(-\inprod{\theta^\star}{\diff_i}))}\right] \diff_i.
\end{align*}
Define a random vector $V \in \real^\numobs$ with independent
components as
\begin{align*}
V_i = \begin{cases} \frac{
    \exp(-\inprod{\theta^\star}{\diff_i})}{1+\exp(-\inprod{\theta^\star}{\diff_i}))} &
  \mbox{w.p. \quad}\frac{
    1}{1+\exp(-\inprod{\theta^\star}{\diff_i}))}\\
\frac{
    -1}{1+\exp(-\inprod{\theta^\star}{\diff_i}))}& \mbox{w.p. \quad} \frac{
    \exp(-\inprod{\theta^\star}{\diff_i})}{1+\exp(-\inprod{\theta^\star}{\diff_i}))}.
\end{cases}
\end{align*}
With this notation, we have $\nabla \loss_\mathcal{D}(\theta^\star) = -
\frac{1}{\numobs } \, \Xmat^\top V$. One can verify that
$\Exs[V] = 0$ and $| V_i| \leq 1$. 

Defining the $\numobs$-dimensional square matrix $M \coloneqq
\frac{1}{ \numobs^2} \diffmx (\Sigma_{\mathcal{D}}+\lambda I)^{-1} \diffmx^\top$, we have $\|{\nabla \loss_\mathcal{D}(\theta^\star)}\|_{(\Sigma_{\mathcal{D}}+\lambda I)^{-1}}^2
=V^\top M V$.
Let the eigenvalue decomposition of $X^\top X$ be $X^\top X= U \Lambda U^\top$. We can bound the trace and operator norm of $M$ as 
\begin{align*}
\mathsf{Tr}(M) &= \frac{1}{n^2}\mathsf{Tr}(U(\Lambda/n+\lambda I)^{-1} U^\top U\Lambda U^\top) \leq  \frac{d}{
  \numobs}  \\
  \mathsf{Tr}(M^2) &= \frac{1}{n^4}\mathsf{Tr}(U(\Lambda/n+\lambda I)^{-1} U^\top U\Lambda U^\top U(\Lambda/n+\lambda I)^{-1} U^\top U\Lambda U^\top) \leq  \frac{d}{
  \numobs^2}  \\
  \|{M}\|_{\mathsf{op}} &= \lambda_{\mathsf{max}}(M)\leq 
\frac{1}{ \numobs},
\end{align*} 
Moreover, since the components of $V$ are independent and of zero mean, and $|V_i| \leq 1$, the variables are
$1$-sub-Gaussian, and hence the Bernstein's inequality for  sub-Gaussian random variables in quadratic form (see e.g. \citet[Theorem 2.1]{hsu2012tail})
implies that  with probability at least $1-\delta$, 
\begin{align*}
    \|{\nabla \loss_\mathcal{D}(\theta^\star)}\|^2_{(\Sigma_{\mathcal{D}}+\lambda I)^{-1}} =V^\top M V \leq C_1\cdot \frac{d+\log(1/\delta)}{n}.
\end{align*}
Here $C_1$ is some universal constant. This gives us 
\begin{align*}
    \strongcon \|{\Delta}\|_{\Sigma_{\mathcal{D}}+\lambda I}^2&\leq \|{\nabla \loss_\mathcal{D}(\theta^\star)}\|_{(\Sigma_{\mathcal{D}}+\lambda I)^{-1}} \;
\|{\Delta}\|_{\Sigma_{\mathcal{D}}+\lambda I} +4\lambda \gamma  B^2\\ 
& \leq \sqrt{C_1\cdot \frac{d+\log(1/\delta)}{ n}}\|{\Delta}\|_{\Sigma_{\mathcal{D}}+\lambda I} +4\lambda \gamma B^2.
\end{align*}
Solving the above inequality gives us that for some constant $C_2$,
\begin{align*}
    \|{\Delta}\|_{\Sigma_{\mathcal{D}}+\lambda I}\leq C_2\cdot \sqrt{ \frac{d+\log(1/\delta)}{\gamma^2 n} +  {\lambda B^2}}.
\end{align*}

\subsection{Proof of Theorem~\ref{thm:LCB_upper}}\label{proof:lcb_pairwise}

\begin{proof}
Let $J'(\pi) = J(\pi) - \inprod{\theta^\star}{v}$. 
    We have
    \begin{align*}
    \mathsf{SubOpt}(\hat \pi_{\mathsf{PE}}) &= 
       J(\pi^\star) - J(\hat \pi_{\mathsf{PE}})  \\
       & =J'(\pi^\star) - J'(\hat\pi_{\mathsf{PE}})  \\
       & =(J'(\pi^\star) -\hat J(\pi^\star)) + (\hat J(\pi^\star) -  \hat J(\hat\pi_{\mathsf{PE}}) )+  ( \hat J(\hat\pi_{\mathsf{PE}}) -J'(\hat\pi_{\mathsf{PE}})).
    \end{align*}
Since $\hat\pi_{\mathsf{PE}}$ is the optimal policy under expected value $J'(\pi)$, we know that  the second difference satisfies $ \hat J(\pi^\star) -  \hat J(\hat\pi_{\mathsf{PE}}) \leq 0 $. For the third difference, we have
\begin{align*}
    \hat J(\hat\pi_{\mathsf{PE}}) -J'(\hat\pi_{\mathsf{PE}}) = \min_{\theta\in \Theta(\hat  \theta_{\mathsf{MLE}},\lambda)}
\mathbb{E}_{s\sim \rho}[\theta^\top (\phi(s, \pi(s))-v)] - \mathbb{E}_{s\sim \rho}[\theta^{\star\top} (\phi(s, \pi(s))-v)].
\end{align*}
From Lemma~\ref{lem:mle_estimation} we know that $\theta^\star\in \Theta(\hat  \theta_{\mathsf{MLE}},\lambda)$ with probability at least $1-\delta$. Thus we know that with probability at least $1-\delta$,        $ \hat J(\hat\pi_{\mathsf{PE}}) -J'(\hat\pi_{\mathsf{PE}})\leq 0$. Now combining everything together and condition on the above event, we have 
\begin{align*}
       \mathsf{SubOpt}(\hat \pi_{\mathsf{PE}})  & \leq J'(\pi^\star) -\hat J(\pi^\star)\\ 
       & =\sup_{\theta\in\Theta(\hat \theta_{\mathsf{MLE}},\lambda)}\mathbb{E}_{s\sim \rho}[(\theta^\star -  \theta)^\top(\phi(s, \pi^\star(s))-v)] \\
       & =\sup_{\theta\in\Theta(\hat \theta_{\mathsf{MLE}},\lambda)}\mathbb{E}_{s\sim \rho}[(\theta^\star -  \hat\theta_{\mathsf{MLE}} +\hat\theta_{\mathsf{MLE}} -\theta)^\top(\phi(s, \pi^\star(s))-v)] \\
       &   = \mathbb{E}_{s\sim \rho}[(\theta^\star -  \hat\theta_{\mathsf{MLE}})^\top(\phi(s, \pi^\star(s))-v)] +\sup_{\theta\in\Theta(\hat \theta_{\mathsf{MLE}},\lambda)}\mathbb{E}_{s\sim \rho}[(\hat\theta_{\mathsf{MLE}} -\theta)^\top(\phi(s, \pi^\star(s))-v)]. 
       \end{align*}
By the definition of $\Theta(\hat \theta_{\mathsf{MLE}}, \lambda)$, we know that for any $\theta\in\Theta(\hat \theta_{\mathsf{MLE}}, \lambda)$, one has $ \mathbb{E}_{s\sim \rho}[(\hat\theta_{\mathsf{MLE}} -\theta)^\top(\phi(s, \pi^\star(s))-v)]\leq C\cdot \sqrt{ \frac{d+\log(1/\delta)}{\gamma^2 n} +  {\lambda B^2}}\cdot \|(\Sigma_{\mathcal{D}}+\lambda I)^{-1/2}\mathbb{E}_{s\sim \rho}[\phi(s, \pi^\star(s))-v]\|_2$.
Furthermore, we know that $ \theta^\star\in\Theta(\hat \theta_{\mathsf{MLE}}, \lambda)$ from Lemma~\ref{lem:mle_estimation}. Altogether we have with probability $1-\delta$ 
\begin{align*}
       \mathsf{SubOpt}(\hat \pi_{\mathsf{PE}}) & \leq 2 C\cdot \sqrt{ \frac{d+\log(1/\delta)}{\gamma^2 n} +  {\lambda B^2}}\cdot \|(\Sigma_{\mathcal{D}}+\lambda I)^{-1/2}\mathbb{E}_{s\sim \rho}[\phi(s, \pi^\star(s))-v]\|_2.
    \end{align*}
    
\end{proof}

\subsection{Proof of Theorem~\ref{thm:fail_mle}}\label{proof:fail_mle}

\begin{proof}
    Consider $4$ actions with parameter $\phi(a_1) = [1,1,0]$, $\phi(a_2) = [1,0,0]$, $\phi(a_3) = [0,0,0]$, $\phi(a_4) = [0,1,0]$. Let the true reward be $\theta^\star = [-1, 0.1,0.9]\in \Theta_B$ with $B=2$.  We query $n-1$ times $a_1, a_2$ and $1$ time $a_2, a_3$. For the single pairwise comparison result $Y_{2>3}$ between $a_2$ and $a_3$, we know that 
    \begin{align*}
        P(Y_{2>3}=1) =\frac{\exp((\phi(a_2)-\phi(a_3))^\top\theta^\star)}{1+\exp((\phi(a_2)-\phi(a_3))^\top\theta^\star)} > 0.26.
    \end{align*}
    Now conditioned on the event that $Y_{2>3}=1$, we know that the MLE aims to find
\begin{align*}
\hat \theta_{\mathsf{MLE}} & = \argmin_{\theta\in\Theta_B}    \ell_\mathcal{D}(\theta), \\ 
\text{where } \ell_\mathcal{D}(\theta) & = -n_{1>2}\cdot \log\left(\frac{\exp((\phi(a_1)-\phi(a_2))^\top\theta)}{1+\exp((\phi(a_1)-\phi(a_2))^\top\theta)}\right)-n_{1<2}\cdot \log\left(\frac{\exp((\phi(a_2)-\phi(a_1))^\top\theta)}{1+\exp((\phi(a_2)-\phi(a_1))^\top\theta)}\right) \\
& \quad - \log\left(\frac{\exp((\phi(a_2)-\phi(a_3))^\top\theta)}{1+\exp((\phi(a_2)-\phi(a_3))^\top\theta)}\right) \\
& = -n_{1>2}\cdot \log\left(\frac{\exp(\theta_2)}{1+\exp(\theta_2)}\right)-n_{1<2}\cdot \log\left(\frac{\exp(-\theta_2)}{1+\exp(-\theta_2)}\right)  - \log\left(\frac{\exp(\theta_1)}{1+\exp(\theta_1)}\right). 
\end{align*}
By concentration of $n_{1>2}$, we know that when $n>500$, with probability at least $0.5$, we have
\begin{align*}
    n_{1<2}>0.45n.
\end{align*}
Under this case, the MLE will satisfy at $\hat \theta_1>0, \hat\theta_2<0.5$. Thus the policy based on MLE estimator will choose  action $a_1$ or $a_2$ instead of  the optimal action $a_4$ under the events above. The expected suboptimality is 
\begin{align*}
        \mathbb{E}[V^\star(s) - V^{\hat \pi_{\mathsf{MLE}}}(s)]\geq 0.26*0.5*1>0.1.
    \end{align*}

On the other hand, one can calculate the coverage as 
\begin{align*}
\|\Sigma_{\mathcal{D}}^{-1/2}\mathbb{E}_{s\sim \rho}[\phi(s, \pi^\star(s))]\|_2 = \frac{n}{n-1}.
\end{align*}
Thus by Theorem~\ref{thm:LCB_upper} we know that pessimistic MLE achieves vanishing error.
\end{proof}

\subsection{Proof of Theorem~\ref{thm:lcb_lower}}\label{proof:lcb_lower}
\begin{proof}
Assume without loss of generality that $d/3$ is some integer.
We set $\mathcal{S} = [d/3]$, $\mathcal{A}=\{a_1, a_2, a_3\}$. For each of the $s, a_i$, we set $\phi(s, a_1) = e_{3s+1}, \phi(s, a_2) = e_{3s+2}, \phi(s, a_3) = 0.$ We set the initial distribution of states as $\rho = \mathsf{Unif}([1,2,\cdots, S])$,  the query times $n(s, a_2, a_3) = n/S\cdot (1-2/\Lambda^2), n(s, a_1, a_3) = n/S\cdot (2/\Lambda^2)$.

Let $v_{-1} = [1/d, 1/d+\Delta, -2/d-\Delta]$, $v_{+1} = [1/d+2\Delta, 1/d+\Delta, -2/d-3\Delta]$. We construct $2^S$ instances, indexed by $\tau \in \{\pm 1\}^S$,  where each $\theta_\tau = [v_{\tau_1}, v_{\tau_2},\cdots, v_{\tau_S}]$. One can see that  $\mathbb{E}[V_{\mathcal{Q}}(\pi^\star) - V^\star_{\mathcal{Q}}(\hat \pi)] = 1/S\cdot \sum_{s\in\mathcal{S}} (r_{\mathcal{Q}}(s, \pi^\star(s)) - r_{\mathcal{Q}}(s, \hat \pi(s))).$ Under each $\theta_\tau$, the optimal policy $\pi(s)$ is either $a_1$ or $a_2$. One can verify that 
$\|\Sigma_{\mathcal{D}}^{-1/2}\mathbb{E}_{s\sim\rho}[\phi(s,\pi^\star(s)])]\|_2\leq \Lambda$ and that $\theta_\tau\in\Theta_B$ with $B=1$ when $d>6$ and 
$\Delta<1/(6d)$.

Furthermore, for any $\theta_\tau, \theta_{\tau'}$ that differs only in the $j$-th coordinate of $\tau$, we have
\begin{align*}
    1/S\cdot  (r_{\mathcal{Q}_{\tau}}(j, \pi^\star(j)) - r_{\mathcal{Q}_{\tau}}(j, \hat \pi(j))+r_{\mathcal{Q}_{\tau'}}(j, \pi^\star(j)) - r_{\mathcal{Q}_{\tau'}}(j, \hat \pi(j)))\geq \Delta/S.
\end{align*}
Thus by Assouad's lemma (see e.g.~\cite{yu1997assouad}), we have 
\begin{align*}
    \inf_{\hat \pi} \sup_{\mathcal{Q}\in \mathsf{CB}(\lambda)} \mathbb{E}[V_{\mathcal{Q}}(\pi^\star) - V^\star_{\mathcal{Q}}(\hat \pi)]& \geq S\cdot \frac{\Delta}{2S} \min_{\tau\sim\tau'}(1-\mathsf{TV(\mathbb{P}_{\theta_{\tau}},\mathbb{P}_{\theta_{\tau'}} )}) \\ 
    & \geq \frac{\Delta}{4} \min_{\tau\sim\tau'}\exp(-D_\mathsf{KL}(\mathbb{P}_{\theta_{\tau}},\mathbb{P}_{\theta_{\tau'}} )).
\end{align*}
Here $\tau\sim\tau'$ refers to any $\tau, \tau'$ that only differs  in one element. And the last inequality is due to the Bretagnolle–Huber inequality~\cite{bretagnolle79}. To bound the KL divergence, we have the following lemma from~\cite{shah2015estimation}:
\begin{lemma}[{\citet{shah2015estimation}}]
    \label{LemKLUpper}
For any pair of quality score vectors $\theta_\tau$ and $\theta_{\tau'}$,
we have
\begin{align}
\label{EqnKLUpper}
\kl{\mprob_{\theta_{\tau}}}{\mprob_{\theta_{\tau}}} & \leq C n (\theta_{\tau} - \theta_{\tau'})^\top \Sigma_{\mathcal{D}}(\theta_{\tau} - \theta_{\tau'}).
\end{align}
\end{lemma}
From the lemma, we have 
\begin{align*}
    \inf_{\hat \pi} \sup_{\mathcal{Q}\in \mathsf{CB}(\lambda)} \mathbb{E}[V_{\mathcal{Q}}(\pi^\star) - V^\star_{\mathcal{Q}}(\hat \pi)]& \geq \frac{\Delta}{2} \min_{\tau\sim\tau'}\exp(-D_\mathsf{KL}(\mathbb{P}_{\theta_{\tau}},\mathbb{P}_{\theta_{\tau'}} )) \\ 
    & \geq \frac{\Delta}{2}\exp(-C n\Delta^2/(S\Lambda^2))
\end{align*}
Taking $\Delta = \Lambda\sqrt{S/n}$ and noting that $S = d/3$ finishes the proof.
\end{proof}

\subsection{Proof of Theorem~\ref{thm:kwise}}
\label{proof:kwise}

This section presents the proof of Theorem~\ref{thm:kwise} for the
setting of $\numchoices$-wise comparisons. 
We first prove the following lemma on the estimation error:
\begin{lemma}
Under the $\numchoices$-wise  PL model, 
 for any $\lambda>0$,
with probability at least $1-\delta$,
\begin{align*}
\|\hat\theta_{\mathsf{MLE}} - \theta^\star\|_{\Sigma_{\mathcal{D}}+\lambda I}\leq C\cdot \sqrt{ \frac{K^4 (d+\log(1/\delta))}{\gamma^2 n} +  {\lambda B^2}}.
\end{align*}
\end{lemma}

Recall that the MLE is given by 
\begin{align*}
\hat \theta_{\mathsf{MLE}} & \in \argmin_{\theta\in\Theta_B}    \ell_\mathcal{D}(\theta), \\ 
\text{where } \ell_\mathcal{D}(\theta) & = - \frac{1}{n}\sum_{i=1}^n \sum_{j=0}^{K-1} \log \left(\frac{\exp(\inprod{\theta}{\phi(s^i, a^i_{\sigma_i(j)})} )}{\sum_{k=j}^{\numchoices-1} \exp(\inprod{\theta}{\phi(s^i, a^i_{\sigma_i(k)})})}\right).
\end{align*}

Our goal is to bound the estimation error 
of the MLE in the squared semi-norm
$\|v\|_{\Sigma_\mathcal{D}+\lambda I}^2 = {v^\top (\Sigma_\mathcal{D}+\lambda I) v}$.

\paragraph{Strong convexity of $\ell$.}

We first show that $\loss_\mathcal{D}$ is strongly convex at $\theta^\star$ with respect to the semi-norm $\|\cdot\|_{\Sigma_{\mathcal{D}}}$,
meaning that there is some constant $\strongcon > 0$ such that
\begin{align}
\loss_\mathcal{D}(\theta^\star + \Delta) - \loss_\mathcal{D}(\theta^\star) - \inprod{\nabla
  \loss_\mathcal{D}(\theta^\star)}{\Delta} & \geq \strongcon \|\Delta\|_{\Sigma_{\mathcal{D}}}^2
\end{align}
for all perturbations $\Delta \in \real^\numitems$ such that $\theta^\star
+ \Delta\in \Theta_B$.

The gradient of the negative log likelihood is
\begin{align*}
\nabla \loss_\mathcal{D}(\theta) &  
= -\frac{1}{\numobs} \sum_{i=1}^{\numobs}
\sum_{j=0}^{K-1}\sum_{k=j}^{K-1}\frac{\exp(\inprod{\theta}{\phi(s^i, a^i_{\sigma_i(k)})})}{\sum_{k'=j}^{K-1} \exp(\inprod{\theta}{\phi(s^i, a^i_{\sigma_i(k')})})}\cdot (\phi(s^i, a^i_{\sigma_i(j)}) - \phi(s^i, a^i_{\sigma_i(k)})).
\end{align*}
The Hessian of the negative log likelihood can be written as
\begin{align*}
&\nabla^2 \loss_\mathcal{D}(\theta) \\
 = &\frac{1}{\numobs} \sum_{i=1}^{\numobs}
\sum_{j=0}^{K-1}\sum_{k=j}^{K-1}\sum_{k'=j}^{K-1}\frac{\exp(\inprod{\theta}{\phi(s^i, a^i_{\sigma_i(k)})+\phi(s^i, a^i_{\sigma_i(k')})})}{2(\sum_{k'=j}^{K-1} \exp(\inprod{\theta}{\phi(s^i, a^i_{\sigma_i(k')})}))^2} \cdot (\phi(s^i, a^i_{\sigma_i(k)}) - \phi(s^i, a^i_{\sigma_i(k')}))(\phi(s^i, a^i_{\sigma_i(k)}) - \phi(s^i, a^i_{\sigma_i(k')}))^\top.
\end{align*} 
Since $\exp(\inprod{\theta}{\phi})\in[\exp(-LB), \exp(LB)]$, we know that the coefficients satisfy
\begin{align*}
\frac{\exp(\inprod{\theta}{\phi(s^i, a^i_{\sigma_i(k)})+\phi(s^i, a^i_{\sigma_i(k')})})}{(\sum_{k'=j}^{K-1} \exp(\inprod{\theta}{\phi(s^i, a^i_{\sigma_i(k')})}))^2} \geq \frac{\exp(-4LB)}{2(K-j)^2}.
\end{align*}
Set $\gamma = \exp(-4LB)/2$.
We can verify that   for any vector $v\in\mathbb{R}^K$, one has
\begin{align*}
    v^\top \nabla^2 \loss_\mathcal{D}(\theta)v  
& \geq \frac{\gamma}{\numobs}  v^\top \left(\sum_{i=1}^{\numobs}
\sum_{j=0}^{\numchoices-1}\frac{1}{(K-j)^2}\sum_{k=j}^{K-1}\sum_{k'=k}^{K-1}(\phi(s^i,a^i_{\sigma_i(k)}) - \phi(s^i,a^i_{\sigma_i(k')}))(\phi(s^i,a^i_{\sigma_i(k)}) - \phi(s^i,a^i_{\sigma_i(k')}))^\top\right)v \\
& \geq \frac{\gamma}{\numobs}  v^\top \left(\sum_{i=1}^{\numobs} \min_{\sigma_i\in\Pi[K]}
\sum_{j=0}^{\numchoices-1}\frac{1}{(K-j)^2}\sum_{k=j}^{K-1}\sum_{k'=k}^{K-1}(\phi(s^i,a^i_{\sigma_i(k)}) - \phi(s^i,a^i_{\sigma_i(k')}))(\phi(s^i,a^i_{\sigma_i(k)}) - \phi(s^i,a^i_{\sigma_i(k')}))^\top\right)v\\
& \geq {\gamma}  v^\top \Sigma_{\mathcal{D}} v \\ 
& = \gamma \|v\|_{\Sigma_{\mathcal{D}}}^2.
\end{align*}
Thus we know that $\loss$ is $\gamma$-strongly convex with respect to the semi-norm $\|\cdot\|_{\Sigma_{\mathcal{D}}}$.

\paragraph{Bounding the estimation error.} Now we aim at bounding the estimation error $\|{\hat\theta_{\mathsf{MLE}} - \theta^\star}\|_{\Sigma_{\mathcal{D}+\lambda I}}$.

Since $\hat\theta_{\mathsf{MLE}}$ is optimal for $\loss_\mathcal{D}$, we have $\loss_\mathcal{D}(\hat\theta_{\mathsf{MLE}}) \leq
\loss_\mathcal{D}(\theta^\star)$.  Defining the error vector $\Delta = \hat\theta_{\mathsf{MLE}} -
\theta^\star$, adding and subtracting the quantity $\inprod{\nabla
  \loss_\mathcal{D}(\theta^\star)}{\Delta}$ yields the bound
\begin{align*}
\loss_\mathcal{D}(\theta^\star + \Delta) - \loss_\mathcal{D}(\theta^\star) - \inprod{\nabla
  \loss_\mathcal{D}(\theta^\star)}{\Delta} & \leq - \inprod{\nabla
  \loss_\mathcal{D}(\theta^\star)}{\Delta}.
\end{align*}
By the $\strongcon$-convexity condition, the left-hand side is lower
bounded by $\strongcon \LAPNORM{\Delta}^2$.  As for the right-hand
side, note that $|\inprod{\nabla
  \loss_\mathcal{D}(\theta^\star)}{\Delta}| \leq \|{\nabla \loss_\mathcal{D}(\theta^\star)}\|_{(\Sigma_{\mathcal{D}}+\lambda I)^{-1}} \;
\|{\Delta}\|_{\Sigma_{\mathcal{D}}+\lambda I}$ for any $\lambda>0$. Altogether we have
\begin{align*}
    \strongcon \LAPNORM{\Delta}^2\leq \|{\nabla \loss_\mathcal{D}(\theta^\star)}\|_{(\Sigma_{\mathcal{D}}+\lambda I)^{-1}} \;
\|{\Delta}\|_{\Sigma_{\mathcal{D}}+\lambda I}. 
\end{align*}

Now we further bound the term $\|{\nabla \loss_\mathcal{D}(\theta^\star)}\|_{(\Sigma_{\mathcal{D}}+\lambda I)^{-1}}$. 
Observe that the gradient takes the form 
\begin{align}\label{eq:gradient_pl}
\nabla \loss_\mathcal{D}(\theta^\star) = -\frac{1}{\numobs} \sum_{i=1}^{\numobs}
\sum_{j=0}^{K-1}\sum_{k=j}^{K-1}\frac{\exp(\inprod{\theta^\star}{\phi(s^i, a^i_{\sigma_i(k)})})}{\sum_{k'=j}^{K-1} \exp(\inprod{\theta^\star}{\phi(s^i, a^i_{\sigma_i(k')})})}\cdot (\phi(s^i, a^i_{\sigma_i(j)}) - \phi(s^i, a^i_{\sigma_i(k)})).
\end{align}
We set 
$x^i_{jk} = \phi(s^i, a_j^i) - \phi(s^i, a_k^i)$. $X\in\mathbb{R}^{(nK(K-1)/2)\times d}$ has the differencing vector 
$x^i_{jk}$ as its $(iK(K-1)/2+k+\sum_{l=K-j+1}^K l )^{th}$ row.  We also define $V^i_{jk}$ be the random variable of the coefficient of $x^i_{jk}$ in Equation (\ref{eq:gradient_pl}) under the PL model, i.e.
conditioned on an arbitrary permutation $\sigma_i$, \begin{align*}
    V^i_{jk} = \begin{cases}
\frac{\exp(\inprod{\theta^\star}{\phi(s^i, a^i_{k})})}{\sum_{k'=\sigma_i^{-1}(j)}^{K-1} \exp(\inprod{\theta^\star}{\phi(s^i, a^i_{\sigma_i(k')})})}, & \text{ if }
 \sigma_i^{-1}(j)<\sigma_i^{-1}(k) \\ 
 -\frac{\exp(\inprod{\theta^\star}{\phi(s^i, a^i_{j})})}{\sum_{k'=\sigma_i^{-1}(k)}^{K-1} \exp(\inprod{\theta^\star}{\phi(s^i, a^i_{\sigma_i(k')})})}, & \text{ otherwise. } 
 \end{cases}
\end{align*}
Here $ \sigma_i^{-1}(j)<\sigma_i^{-1}(k)$ means that the $j$-th item ranks higher than the $k$-th item. 
Let $\tilde V_i\in\mathbb{R}^{K(K-1)/2} $ be the concatenated random vector of $\{V^i_{jk}\}_{0\leq j<k\leq K-1}$, $V\in\mathbb{R}^{nK(K-1)/2}$ be the concatenated random vector of $\{\tilde V_i\}_{i=1}^n$. We know that $\tilde V_i$ and $\tilde V_j$ are independent for each $i\neq j$ due to the independent sampling procedure. We can also verify that the mean of $\tilde V_i$ is $0$, the proof of which is deferred to the end of this section. Furthermore, since under any permutation, the sum of absolute value of each element in  $\tilde V_i$ is at most $K$, we know that  $\tilde V_i$  is sub-Gaussian with parameter $K$. Thus we know that  $V$ is also sub-Gaussian with mean ${0}$ and parameter $K$. Now we know that the term  $\|{\nabla \loss_\mathcal{D}(\theta^\star)}\|_{(\Sigma_{\mathcal{D}}+\lambda I)^{-1}}^2$ can be written as
\begin{align*}
\|{\nabla \loss_\mathcal{D}(\theta^\star)}\|_{(\Sigma_{\mathcal{D}}+\lambda I)^{-1}}^2 = \frac{1}{n^2} V^\top 
X  (\Sigma_{\mathcal{D}}+\lambda I)^{-1} X^\top V.
\end{align*}
Let $M = \frac{K^2}{n} I$. One can verify that $M \succeq \frac{1}{n^2}X  (\Sigma_{\mathcal{D}}+\lambda I)^{-1} X^\top $ almost surely since $\lambda_{\mathsf{max}}(X  (\Sigma_{\mathcal{D}}+\lambda I)^{-1} X^\top/n^2)\leq K^2/ n$. Thus we can upper bound the original term as
\begin{align*}
\|{\nabla \loss_\mathcal{D}(\theta^\star)}\|_{(\Sigma_{\mathcal{D}}+\lambda I)^{-1}}^2 \leq \frac{K^2}{n} \|V\|_2^2. 
\end{align*}
By Bernstein's inequality for  sub-Gaussian random variables in quadratic form (see e.g. \citet[Theorem 2.1]{hsu2012tail}), we know that with probability at least $1-\delta$,
\begin{align*}
    \|V\|_2^2\leq  C K^2\cdot (d+\log(1/\delta)).
\end{align*}
Thus altogether, we have \begin{align*}
    \strongcon \LAPNORM{\Delta}^2\leq  \sqrt{ \frac{C K^4 \cdot (d+\log(1/\delta))}{n}}
\|{\Delta}\|_{\Sigma_{\mathcal{D}}+\lambda I}. 
\end{align*}
Similar to the pairwise comparison analysis in Appendix~\ref{proof:mle_estimation}, we can derive that with probability at least $1-\delta$,
\begin{align*}
    \|\hat\theta_{\mathsf{MLE}}-\theta^\star\|_{\Sigma_{\mathcal{D}}+\lambda I}\leq C\cdot \sqrt{\frac{K^4 (d+\log(1/\delta))}{n} + \lambda B^2}.
\end{align*}

The rest of the proof on the sub-optimality upper bound follows the same argument as Theorem~\ref{thm:LCB_upper}. 

Lastly, we verify that the mean of $\tilde V_i$ is $0$. For any fixed $j, k\in[K]$, let $\mathcal{P}$ be the ordered set of all elements which are ranked higher than both $j$ and $k$. 
Now conditioned on $\mathcal{P}$, we have

\begin{resizealign}
    \mathbb{E}[V^i_{jk} \mid \mathcal{P}] & = \mathbb{P}(j \text{ follows }\mathcal{P}\mid\mathcal{P}) \cdot \frac{\exp(\inprod{\theta^\star}{\phi(s^i, a^i_{k})})}{\sum_{k'\in \bar{ \mathcal{P}}} \exp(\inprod{\theta^\star}{\phi(s^i, a^i_{k'})})} - \mathbb{P}(k \text{ follows }\mathcal{P}\mid\mathcal{P}) \cdot \frac{\exp(\inprod{\theta^\star}{\phi(s^i, a^i_{j})})}{\sum_{k'\in \bar{ \mathcal{P}}} \exp(\inprod{\theta^\star}{\phi(s^i, a^i_{k'})})}  \nonumber
    \\ 
    & = \frac{1}{\sum_{k'\in \bar{ \mathcal{P}}} \exp(\inprod{\theta^\star}{\phi(s^i, a^i_{k'})})} \cdot \left(\frac{\exp(\inprod{\theta^\star}{\phi(s^i, a^i_{j})})\exp(\inprod{\theta^\star}{\phi(s^i, a^i_{k})})-\exp(\inprod{\theta^\star}{\phi(s^i, a^i_{j})})\exp(\inprod{\theta^\star}{\phi(s^i, a^i_{k})})}{\exp(\inprod{\theta^\star}{\phi(s^i, a^i_{j})})+\exp(\inprod{\theta^\star}{\phi(s^i, a^i_{k})})} \right) \nonumber \\
    & = 0.\nonumber 
\end{resizealign}

Here the second equality uses the fact that $j$ follows $\mathcal{P}$ is equivalent to the event that $j$ is larger than $k$ and either $j, k$ is the largest among $\bar{\mathcal{P}}$. 
Taking expectation over $\mathcal{P}$ gives us that $\mathbb{E}[V^i_{jk}] =0$.

\subsection{Proof of Theorem~\ref{thm:kwise_pair}}
\label{proof:kwise_pair}

This section presents the proof of Theorem~\ref{thm:kwise_pair} for the
setting of $\numchoices$-wise comparisons.  We first prove the following lemma on the estimation error.
\begin{lemma}
    Under the $\numchoices$-wise  PL model, 
 for any $\lambda>0$,
with probability at least $1-\delta$,
\begin{align*}
\|\hat \theta_{\mathsf{MLE}_2} - \theta^\star\|_{\Sigma_{\mathcal{D}}+\lambda I}\leq C\cdot \sqrt{ \frac{d+\log(1/\delta)}{\gamma^2 n} +  {\lambda B^2}}.
\end{align*}
\end{lemma}

Recall that the pairwise compairson based estimator is given by 
\begin{align*}
\hat \theta_{\mathsf{MLE}_2} & \in \argmin_{\theta\in\Theta_B}    \ell_\mathcal{D}(\theta), \\ 
\text{where } \ell_\mathcal{D}(\theta) & = - \frac{1}{n}\sum_{i=1}^n \sum_{j=0}^{K-1} \sum_{k=j+1}^{K-1}\log \left(\frac{\exp(\inprod{\theta}{\phi(s^i, a^i_{\sigma_i(j)})} )}{\exp(\inprod{\theta}{\phi(s^i, a^i_{\sigma_i(j)})} )+\exp(\inprod{\theta}{\phi(s^i, a^i_{\sigma_i(k)})})}\right).
\end{align*}

Our goal is to bound the estimation error 
of the MLE in the squared semi-norm
$\|v\|_{\Sigma_\mathcal{D}+\lambda I}^2 = {v^\top (\Sigma_\mathcal{D}+\lambda I) v}$.

\paragraph{Strong convexity of $\ell$.}
Let $x^i_{jk} = \phi(s^i,a^i_j) - \phi(s^i, a_k^i)$. 
The gradient of the negative log likelihood is
\begin{align*}
\nabla \loss_\mathcal{D}(\theta) & 
= -\frac{1}{\numobs} \sum_{i=1}^{\numobs}
\sum_{j=0}^{K-1}\sum_{k=j+1}^{K-1}\frac{\exp(-\inprod{\theta}{x^i_{\sigma_i(j)\sigma_i(k)})}}{1+\exp(-\inprod{\theta}{x^i_{\sigma_i(j)\sigma_i(k)})}}\cdot x^i_{\sigma_i(j)\sigma_i(k)}.
\end{align*}
The Hessian of the negative log likelihood can be written as
\begin{align*}
\nabla^2 \loss_\mathcal{D}(\theta) 
 = \frac{1}{\numobs} \sum_{i=1}^{\numobs}
\sum_{j=0}^{K-1}\sum_{k=j}^{K-1}\frac{\exp(-\inprod{\theta}{x^i_{\sigma_i(j)\sigma_i(k)})}}{(1+\exp(-\inprod{\theta}{x^i_{\sigma_i(j)\sigma_i(k)})})^2} \cdot x^i_{\sigma_i(j)\sigma_i(k)}x^{i\top}_{\sigma_i(j)\sigma_i(k)}.
\end{align*}
 
Since $\exp(\inprod{\theta}{x^i_{\sigma_i(j)\sigma_i(k)}})\in[\exp(-2LB), \exp(2LB)]$, we know that the coefficients satisfy
\begin{align*}
\frac{\exp(-\inprod{\theta}{x^i_{\sigma_i(j)\sigma_i(k)})}}{(1+\exp(-\inprod{\theta}{x^i_{\sigma_i(j)\sigma_i(k)})})^2} \geq \frac{1}{2+\exp(2LB)+\exp(-2LB)}.
\end{align*}
Set $\gamma = \frac{1}{2+\exp(2LB)+\exp(-2LB)}$.
We can verify that   for any vector $v\in\mathbb{R}^K$, one has
\begin{align*}
    v^\top \nabla^2 \loss_\mathcal{D}(\theta)v 
& \geq \frac{\gamma}{\numobs}  v^\top \left(\sum_{i=1}^{\numobs}
\sum_{j=0}^{\numchoices-1}\sum_{k=j+1}^{K-1}x^i_{\sigma_i(j)\sigma_i(k)}x^{i\top}_{\sigma_i(j)\sigma_i(k)}\right)v \\
& = \frac{\gamma}{\numobs}  v^\top \left(\sum_{i=1}^{\numobs}
\sum_{j=0}^{\numchoices-1}\sum_{k=j+1}^{K-1}x^i_{jk}x^{i\top}_{jk}\right)v \\
& = {\gamma} K(K-1)  v^\top \Sigma_{\mathcal{D}} v /2 \\ 
& = \gamma  K(K-1)\|v\|_{\Sigma_{\mathcal{D}}}^2/2.
\end{align*}
Thus we know that $\loss$ is $\gamma$-strongly convex with respect to the semi-norm $\|\cdot\|_{\Sigma_{\mathcal{D}}}$.

\paragraph{Bounding the estimation error.} Now we aim at bounding the estimation error $\|{\hat \theta_{\mathsf{MLE}_2} - \theta^\star}\|_{\Sigma_{\mathcal{D}+\lambda I}}$.

Since $\hat \theta_{\mathsf{MLE}_2}$ is optimal for $\loss_\mathcal{D}$, we have $\loss_\mathcal{D}(\hat \theta_{\mathsf{MLE}_2}) \leq
\loss_\mathcal{D}(\theta^\star)$.  Defining the error vector $\Delta = \hat \theta_{\mathsf{MLE}_2} -
\theta^\star$, adding and subtracting the quantity $\inprod{\nabla
  \loss_\mathcal{D}(\theta^\star)}{\Delta}$ yields the bound
\begin{align*}
\loss_\mathcal{D}(\theta^\star + \Delta) - \loss_\mathcal{D}(\theta^\star) - \inprod{\nabla
  \loss_\mathcal{D}(\theta^\star)}{\Delta} & \leq - \inprod{\nabla
  \loss_\mathcal{D}(\theta^\star)}{\Delta}.
\end{align*}
By the $\strongcon$-convexity condition, the left-hand side is lower
bounded by $\strongcon K(K-1)\LAPNORM{\Delta}^2/2$.  As for the right-hand
side, note that $|\inprod{\nabla
  \loss_\mathcal{D}(\theta^\star)}{\Delta}| \leq  \|{\nabla \loss_\mathcal{D}(\theta^\star)}\|_{(\Sigma_{\mathcal{D}}+\lambda I)^{-1}} \;
\|{\Delta}\|_{\Sigma_{\mathcal{D}}+\lambda I}$ for any $\lambda>0$. Altogether we have
\begin{align*}
    \strongcon \LAPNORM{\Delta}^2\leq 2\|{\nabla \loss_\mathcal{D}(\theta^\star)}\|_{(\Sigma_{\mathcal{D}}+\lambda I)^{-1}} \;
\|{\Delta}\|_{\Sigma_{\mathcal{D}}+\lambda I}/K(K-1). 
\end{align*}

Now we further bound the term $\|{\nabla \loss_\mathcal{D}(\theta^\star)}\|_{(\Sigma_{\mathcal{D}}+\lambda I)^{-1}}$. 
Observe that the gradient takes the form 
\begin{align}\label{eq:gradient_pl_2}
\nabla \loss_\mathcal{D}(\theta^\star) = -\frac{1}{\numobs} \sum_{i=1}^{\numobs}
\sum_{j=0}^{K-1}\sum_{k=j+1}^{K-1}\frac{\exp(-\inprod{\theta}{x^i_{\sigma_i(j)\sigma_i(k)})}}{1+\exp(-\inprod{\theta}{x^i_{\sigma_i(j)\sigma_i(k)})}}\cdot x^i_{\sigma_i(j)\sigma_i(k)}.
\end{align}
We  set $X\in\mathbb{R}^{(nK(K-1)/2)\times d}$ with the    differencing vector 
$x^i_{jk}$ as its $(iK(K-1)/2+k+\sum_{l=K-j+1}^K l )^{th}$ row.  We also define $V^i_{jk}$ be the random variable of the coefficient of $x^i_{jk}$ in Equation (\ref{eq:gradient_pl_2}) under the PL model, i.e.
conditioned on an arbitrary permutation $\sigma_i$, \begin{align*}
    V^i_{jk} = \begin{cases}
\frac{\exp(-\inprod{\theta}{x^i_{jk})}}{1+\exp(-\inprod{\theta}{x^i_{jk})}}, & \text{ if }
 \sigma_i^{-1}(j)<\sigma_i^{-1}(k) \\ 
 -\frac{1}{1+\exp(-\inprod{\theta}{x^i_{jk})}}, & \text{ otherwise. } 
 \end{cases}
\end{align*}
Let $\tilde V_i\in\mathbb{R}^{K(K-1)/2} $ be the concatenated random vector of $\{V^i_{jk}\}_{0\leq j<k\leq K-1}$, $V\in\mathbb{R}^{nK(K-1)/2}$ be the concatenated random vector of $\{\tilde V_i\}_{i=1}^n$. We know that $\tilde V_i$ is independent for each $i$, and that $V$ is sub-Gaussian with mean $\mathbf{0}$ and parameter $\sqrt{K(K-1)/2}$ since the PL model reduces to BTL model when considering pairwise comparisons. Now we know that the term  $\|{\nabla \loss_\mathcal{D}(\theta^\star)}\|_{(\Sigma_{\mathcal{D}}+\lambda I)^{-1}}^2$ can be written as
\begin{align*}
\|{\nabla \loss_\mathcal{D}(\theta^\star)}\|_{(\Sigma_{\mathcal{D}}+\lambda I)^{-1}}^2 = \frac{1}{n^2} V^\top 
X  (\Sigma_{\mathcal{D}}+\lambda I)^{-1} X^\top V.
\end{align*}
Let $M = \frac{K^2}{n} I$. One can verify that $M \succeq \frac{1}{n^2}X  (\Sigma_{\mathcal{D}}+\lambda I)^{-1} X^\top $ almost surely since $\lambda_{\mathsf{max}}(X  (\Sigma_{\mathcal{D}}+\lambda I)^{-1} X^\top/n^2)\leq K^2/ n$. Thus we can upper bound the original term as
\begin{align*}
\|{\nabla \loss_\mathcal{D}(\theta^\star)}\|_{(\Sigma_{\mathcal{D}}+\lambda I)^{-1}}^2 \leq \frac{K^2}{n} \|V\|_2^2. 
\end{align*}

By Bernstein's inequality for  sub-Gaussian random variables in quadratic form (see e.g. \citet[Theorem 2.1]{hsu2012tail}), we know that with probability at least $1-\delta$,
\begin{align*}
    \|V\|_2^2\leq  C K(K-1)\cdot (d+\log(1/\delta)).
\end{align*}
Thus altogether, we have \begin{align*}
    \strongcon \LAPNORM{\Delta}^2\leq  \sqrt{ \frac{C  \cdot (d+\log(1/\delta))}{n}}
\|{\Delta}\|_{\Sigma_{\mathcal{D}}+\lambda I}. 
\end{align*}
Similar to the pairwise comparison, we can derive that with probability at least $1-\delta$,
\begin{align*}
    \|\hat \theta_{\mathsf{MLE}_2}-\theta^\star\|_{\Sigma_{\mathcal{D}}+\lambda I}\leq C\cdot \sqrt{\frac{ d+\log(1/\delta)}{n} + \lambda B^2}.
\end{align*}
The rest of the proof on the sub-optimality upper bound follows the same argument as Theorem~\ref{thm:LCB_upper}.

\subsection{Proof of Lemma~\ref{lem:mle_estimation_rl}}\label{proof:mle_estimation_rl_traj}

Recall that the MLE is given by 
\begin{align*}
\hat \theta_{\mathsf{MLE}} & \in \argmin_{\theta\in\Theta_B}    \ell_\mathcal{D}(\theta), \\ 
\text{where } \ell_\mathcal{D}(\theta) & = -\sum_{i=1}^n \log \Big(1(y^i=1)\cdot \frac{\exp(\sum_{h=1}^H r_{\theta}(s^i_h, a_{h}^i))}{\exp(\sum_{h=1}^H r_{\theta}(s^i_h, a_{h}^i))+\exp(\sum_{h=1}^H r_{\theta}(s_h^{i\prime}, a_h^{i\prime}))} \\ 
& \qquad +1(y^i=0)\cdot \frac{\exp(\sum_{h=1}^H r_{\theta}(s_h^{i\prime}, a_h^{i\prime}))}{\exp(\sum_{h=1}^H r_{\theta}(s^i_h, a_{h}^i))+\exp(\sum_{h=1}^H r_{\theta}(s_h^{i\prime}, a_h^{i\prime}))}\Big) \\ 
& = -\sum_{i=1}^n \log \Big(1(y^i=1)\cdot \frac{1}{\exp(-\sum_{h=1}^H (r_{\theta}(s^i_h, a_{h}^i)-r_{\theta}(s_h^{i\prime}, a_h^{i\prime})))+1} \\ 
& \qquad +1(y^i=0)\cdot \frac{1}{\exp(\sum_{h=1}^H (r_{\theta}(s^i_h, a_{h}^i)-r_{\theta}(s_h^{i\prime}, a_h^{i\prime})))+1}\Big) \\
& =  -\sum_{i=1}^n \log \Big(1(y^i=1)\cdot \frac{1}{\exp(-\inprod{\theta}{\sum_{h=1}^H (\phi(s^i_h, a_{h}^i)-\phi(s_h^{i\prime}, a_h^{i\prime}))})+1} \\ 
& \qquad +1(y^i=0)\cdot \frac{1}{\exp(\inprod{\theta}{\sum_{h=1}^H (\phi(s^i_h, a_{h}^i)-\phi(s_h^{i\prime}, a_h^{i\prime}))})+1}\Big)
\end{align*}
To simplify the notation, we let $x_i = \sum_{h=1}^H (\phi(s^i_h, a_{h}^i)-\phi(s_h^{i\prime}, a_{h}^{i\prime}))$. 
Our goal is to bound the estimation error 
of the MLE in the squared semi-norm
$\|v\|_{\Sigma_\mathcal{D}+\lambda I}^2 = {v^\top (\Sigma_\mathcal{D}+\lambda I) v}$.

\paragraph{Strong convexity of $\ell$.}

We first show that $\loss_\mathcal{D}$ is strongly convex at $\theta^\star$ with respect to the semi-norm $\|\cdot\|_{\Sigma_{\mathcal{D}}}$,
meaning that there is some constant $\strongcon > 0$ such that
\begin{align}
\loss_\mathcal{D}(\theta^\star + \Delta) - \loss_\mathcal{D}(\theta^\star) - \inprod{\nabla
  \loss_\mathcal{D}(\theta^\star)}{\Delta} & \geq \strongcon \|\Delta\|_{\Sigma_{\mathcal{D}}}^2
\end{align}
for all perturbations $\Delta \in \real^\numitems$ such that $\theta^\star
+ \Delta\in \Theta_B$.

One can directly calculate the Hessian of $\ell$ as
\begin{align*}
\nabla^2 \loss_\mathcal{D}(\theta) = \frac{1}{\numobs}
\sum_{i=1}^{\numobs} \left(1(y^i=1)\cdot\frac{\exp(-\inprod{\theta}{\diff_i})}{(\exp(-\inprod{\theta}{\diff_i})+1)^2} +1(y^i=0)\cdot\frac{\exp(\inprod{\theta}{\diff_i})}{(\exp(\inprod{\theta}{\diff_i})+1)^2} \right)\cdot \diff_i \diff_i^\top,
\end{align*}
Observe that  $\inprod{\theta}{\diff_i}\in[-2HLB,2HLB]$, we have
\begin{align*}
 v^\top \nabla^2 \loss_\mathcal{D} (\theta) v \geq \frac{\strongcon}{\numobs
   } \|X v\|_2^2 \qquad \mbox{for all $v$},
\end{align*}
where $\strongcon = 1/(2+\exp(-2HLB)+\exp(2HLB))$, $\Xmat \in \real^{\numobs \times \numitems}$ has the
differencing vector $x_i \in \real^\numitems$ as its $i^{th}$ row.

Thus, if we introduce the error vector $\Delta \coloneqq \hat\theta_{\mathsf{MLE}} -
\theta^\star$, then we may conclude that
\begin{align*}
\loss_\mathcal{D}(\theta^\star + \Delta) - \loss_\mathcal{D}( \theta^\star ) - \inprod{\nabla \loss_\mathcal{D}
  (\theta^\star)}{ \Delta} & \geq \frac{\strongcon}{\numobs }
\|X \Delta\|_2^2 \; = \; {\strongcon}{}
\|{\Delta}\|_{\Sigma_{\mathcal{D}}}^2,
\end{align*}
showing that $\loss_\mathcal{D}$ is strongly convex around $\theta^\star$ with
parameter ${\strongcon}$.

\paragraph{Bounding the estimation error.} Now we aim at bounding the estimation error $\LAPNORM{\hat\theta_{\mathsf{MLE}} - \theta^\star}$.

Since $\hat\theta_{\mathsf{MLE}}$ is optimal for $\loss_\mathcal{D}$, we have $\loss_\mathcal{D}(\hat\theta_{\mathsf{MLE}}) \leq
\loss_\mathcal{D}(\theta^\star)$.  Defining the error vector $\Delta = \hat\theta_{\mathsf{MLE}} -
\theta^\star$, adding and subtracting the quantity $\inprod{\nabla
  \loss_\mathcal{D}(\theta^\star)}{\Delta}$ yields the bound
\begin{align*}
\loss_\mathcal{D}(\theta^\star + \Delta) - \loss_\mathcal{D}(\theta^\star) - \inprod{\nabla
  \loss_\mathcal{D}(\theta^\star)}{\Delta} & \leq - \inprod{\nabla
  \loss_\mathcal{D}(\theta^\star)}{\Delta}.
\end{align*}
By the $\strongcon$-convexity condition, the left-hand side is lower
bounded by $\strongcon \LAPNORM{\Delta}^2$.  As for the right-hand
side, note that $|\inprod{\nabla
  \loss_\mathcal{D}(\theta^\star)}{\Delta}| \leq \|{\nabla \loss_\mathcal{D}(\theta^\star)}\|_{(\Sigma_{\mathcal{D}}+\lambda I)^{-1}} \;
\|{\Delta}\|_{\Sigma_{\mathcal{D}}+\lambda I}$ for any $\lambda>0$. Altogether we have
\begin{align*}
    \strongcon \LAPNORM{\Delta}^2\leq \|{\nabla \loss_\mathcal{D}(\theta^\star)}\|_{(\Sigma_{\mathcal{D}}+\lambda I)^{-1}} \;
\|{\Delta}\|_{\Sigma_{\mathcal{D}}+\lambda I}. 
\end{align*}

Now we further bound the term $\|{\nabla \loss_\mathcal{D}(\theta^\star)}\|_{(\Sigma_{\mathcal{D}}+\lambda I)^{-1}}$. 
Observe that the gradient takes the form
\begin{align*}
\nabla \loss_\mathcal{D} (\theta^\star) & = \frac{-1}{\numobs }
\sum_{i=1}^{\numobs} \left[ \Ind[\obs^i = 1] \frac{
    \exp(-\inprod{\theta^\star}{\diff_i})}{1+\exp(-\inprod{\theta^\star}{\diff_i}))} - \Ind[\obs^i=0] \frac{
    1}{1+\exp(-\inprod{\theta^\star}{\diff_i}))}\right] \diff_i.
\end{align*}
Define a random vector $V \in \real^\numobs$ with independent
components as
\begin{align*}
V_i = \begin{cases} \frac{
    \exp(-\inprod{\theta^\star}{\diff_i})}{1+\exp(-\inprod{\theta^\star}{\diff_i}))} &
  \mbox{w.p. \quad}\frac{
    1}{1+\exp(-\inprod{\theta^\star}{\diff_i}))}\\
\frac{
    -1}{1+\exp(-\inprod{\theta^\star}{\diff_i}))}& \mbox{w.p. \quad} \frac{
    \exp(-\inprod{\theta^\star}{\diff_i})}{1+\exp(-\inprod{\theta^\star}{\diff_i}))}.
\end{cases}
\end{align*}
With this notation, we have $\nabla \loss_\mathcal{D}(\theta^\star) = -
\frac{1}{\numobs } \, \Xmat^\top V$. One can verify that
$\Exs[V] = 0$ and $| V_i| \leq 1$. 

Defining the $\numobs$-dimensional square matrix $M \coloneqq
\frac{1}{ \numobs^2} \diffmx (\Sigma_{\mathcal{D}}+\lambda I)^{-1} \diffmx^\top$, we have $\|{\nabla \loss_\mathcal{D}(\theta^\star)}\|_{(\Sigma_{\mathcal{D}}+\lambda I)^{-1}}
=V^\top M V$.
Let the eigenvalue decomposition of $XX^\top$ be $XX^\top = U \Lambda U^\top$. We can bound the trace and operator norm of $M$ as 
\begin{align*}
\mathsf{Tr}(M) &= \frac{1}{n^2}\mathsf{Tr}(U(\Lambda/n+\lambda I)^{-1} U^\top U\Lambda U^\top) \leq  \frac{d}{
  \numobs}  \\
  \|M\|_{\mathsf{op}} &= \lambda_{\mathsf{max}}(M)\leq 
\frac{1}{ \numobs},
\end{align*} 
Moreover, since the components of $V$ are independent and of zero mean, and $|V_i| \leq 1$, the variables are
$1$-sub-Gaussian, and hence the Bernstein's inequality for  sub-Gaussian random variables in quadratic form (see e.g. \citet[Theorem 2.1]{hsu2012tail})
implies that  with probability at least $1-\delta$, 
\begin{align*}
    \|{\nabla \loss_\mathcal{D}(\theta^\star)}\|^2_{(\Sigma_{\mathcal{D}}+\lambda I)^{-1}} =V^\top M V \leq C_1\cdot \frac{d+\log(1/\delta)}{n}.
\end{align*}
Here $C_1$ is some universal constant. This gives us 
\begin{align*}
    \strongcon \|{\Delta}\|_{\Sigma_{\mathcal{D}}+\lambda I}^2&\leq \|{\nabla \loss_\mathcal{D}(\theta^\star)}\|_{(\Sigma_{\mathcal{D}}+\lambda I)^{-1}} \;
\|{\Delta}\|_{\Sigma_{\mathcal{D}}+\lambda I} +4\lambda \gamma  B^2\\ 
& \leq \sqrt{C_1\cdot \frac{d+\log(1/\delta)}{ n}}\|{\Delta}\|_{\Sigma_{\mathcal{D}}+\lambda I} +4\lambda \gamma B^2.
\end{align*}
Solving the above inequality gives us that for some constant $C_2$,
\begin{align*}
    \|{\Delta}\|_{\Sigma_{\mathcal{D}}+\lambda I}\leq C_2\cdot \sqrt{ \frac{d+\log(1/\delta)}{\gamma^2 n} +  {\lambda B^2}}.
\end{align*}
\subsection{Proof of Theorem~\ref{thm:LCB_upper_RL_traj}}\label{proof:LCB_upper_rl_traj}
\begin{proof}
    From Lemma~\ref{lem:mle_estimation_rl}, we know that with probability at least $1-\delta$,
    \begin{align*}
\|\hat\theta_{\mathsf{MLE}} - \theta^\star\|_{\Sigma_{\mathcal{D}}+\lambda I}\leq C\cdot \sqrt{ \frac{d+\log(1/\delta)}{\gamma^2 n} +  {\lambda B^2}}.
\end{align*}
Let $J'(\pi) = J(\pi) - H\inprod{\theta^\star}{v}$. 
    We have
    \begin{align*}
    \mathsf{SubOpt}(\hat \pi_{\mathsf{PE}}) &= 
       J(\pi^\star) - J(\hat \pi_{\mathsf{PE}})  \\
       & =J'(\pi^\star) - J'(\hat\pi_{\mathsf{PE}})  \\
       & =(J'(\pi^\star) -\hat J(\pi^\star)) + (\hat J(\pi^\star) -  \hat J(\hat\pi_{\mathsf{PE}}) )+  ( \hat J(\hat\pi_{\mathsf{PE}}) -J'(\hat\pi_{\mathsf{PE}})).
    \end{align*}
Since $\hat\pi_{\mathsf{PE}}$ is the optimal policy under expected value $\hat J(\pi)$, we know that  the second difference satisfies $ \hat J(\pi^\star) -  \hat J(\hat\pi_{\mathsf{PE}}) \leq 0 $. For the third difference, we have
\begin{align*}
    \hat J(\hat\pi_{\mathsf{PE}}) -J'(\hat\pi_{\mathsf{PE}}) = \mathbb{E}_{s\sim d^{\hat\pi_{\mathsf{PE}}}}[\hat r(s, \hat \pi_{\mathsf{PE}}(s)) - r(s, \hat\pi_{\mathsf{PE}}(s))].
\end{align*}
From Lemma~\ref{lem:mle_estimation_rl} we know that $\theta^\star\in \Theta(\hat  \theta_{\mathsf{MLE}},\lambda)$ with probability at least $1-\delta$. Thus we know that with probability at least $1-\delta$,        $ \hat J(\hat\pi_{\mathsf{PE}}) -J'(\hat\pi_{\mathsf{PE}})\leq 0$. Now combining everything together, we have 
\begin{align*}
       \mathsf{SubOpt}(\hat \pi_{\mathsf{PE}})  & \leq J'(\pi^\star) -\hat J(\pi^\star)\\ 
       & =\sup_{\theta\in\Theta(\hat \theta_{\mathsf{MLE}},\lambda)}\mathbb{E}_{s\sim d^{\pi^\star}}[(\theta^\star -  \theta)^\top(\phi(s, \pi^\star(s))-v)] \\
       & =\sup_{\theta\in\Theta(\hat \theta_{\mathsf{MLE}},\lambda)}\mathbb{E}_{s\sim d^{\pi^\star}}[(\theta^\star -  \hat\theta_{\mathsf{MLE}} +\hat\theta_{\mathsf{MLE}} -\theta)^\top(\phi(s, \pi^\star(s))-v)] \\
       &   = \mathbb{E}_{s\sim d^{\pi^\star}}[(\theta^\star -  \hat\theta_{\mathsf{MLE}})^\top(\phi(s, \pi^\star(s))-v)] +\sup_{\theta\in\Theta(\hat \theta_{\mathsf{MLE}},\lambda)}\mathbb{E}_{s\sim d^{\pi^\star}}[(\hat\theta_{\mathsf{MLE}} -\theta)^\top(\phi(s, \pi^\star(s))-v)]. 
       \end{align*}
By the definition of $\Theta(\hat \theta_{\mathsf{MLE}}, \lambda)$, we know that for any $\theta\in\Theta(\hat \theta_{\mathsf{MLE}}, \lambda)$, one has $ \mathbb{E}_{s\sim d^{\pi^\star}}[(\hat\theta_{\mathsf{MLE}} -\theta)^\top(\phi(s, \pi^\star(s))-v)]\leq C\cdot \sqrt{ \frac{d+\log(1/\delta)}{\gamma^2 n} +  {\lambda B^2}}\cdot \|(\Sigma_{\mathcal{D}}+\lambda I)^{-1/2}\mathbb{E}_{s\sim d^{\pi^\star}}[\phi(s, \pi^\star(s))-v]\|_2$.
Furthermore, we know that $\hat \theta^\star\in\Theta(\hat \theta_{\mathsf{MLE}}, \lambda)$ from Lemma~\ref{lem:mle_estimation_rl}. Altogether we have with probability $1-2\delta$ 
\begin{align*}
       \mathsf{SubOpt}(\hat \pi_{\mathsf{PE}}) & \leq 2 C\cdot \sqrt{ \frac{d+\log(1/\delta)}{\gamma^2 n} +  {\lambda B^2}}\cdot \|(\Sigma_{\mathcal{D}}+\lambda I)^{-1/2}\mathbb{E}_{s\sim d^{\pi^\star}}[\phi(s, \pi^\star(s))-v]\|_2.
    \end{align*}
\end{proof}
\subsection{Proof of Theorem~\ref{thm:LCB_upper_IRL}}\label{proof:LCB_upper_IRL}
\begin{proof}
Here We mainly prove Lemma~\ref{lem:mle_estimation_IRL}, since Theorem~\ref{thm:LCB_upper_IRL} is a direct corollary when combined with the proof in Theorem~\ref{thm:LCB_upper_RL_traj}.

Our goal is to bound the estimation error 
of the MLE in the squared semi-norm
$\|v\|_{\Sigma_\mathcal{D}+\lambda I}^2 = {v^\top (\Sigma_\mathcal{D}+\lambda I) v}$.

\paragraph{Strong convexity of $\ell$.}

We first show that $\loss_\mathcal{D}$ is strongly convex at $\theta^\star$ with respect to the semi-norm $\|\cdot\|_{\Sigma_{\mathcal{D}}}$,
meaning that there is some constant $\strongcon > 0$ such that
\begin{align}
\loss_\mathcal{D}(\theta^\star + \Delta) - \loss_\mathcal{D}(\theta^\star) - \inprod{\nabla
  \loss_\mathcal{D}(\theta^\star)}{\Delta} & \geq \strongcon \|\Delta\|_{\Sigma_{\mathcal{D}}}^2
\end{align}
for all perturbations $\Delta \in \real^\numitems$ such that $\theta^\star
+ \Delta\in \Theta_B$.

The gradient of the negative log likelihood is
\begin{align*}
\nabla \loss_\mathcal{D}(\theta) & 
= -\frac{1}{\numobs} \sum_{i=1}^{\numobs}
\sum_{\tau'\in\mathcal{T}(s_0^i)}  \frac{\exp(\sum_{h=0}^H \inprod{\theta}{\phi(s_h', a_h')})}{\sum_{\tau''\in\mathcal{T}(s_0^i)} \exp(\sum_{h=0}^H \inprod{\theta}{\phi(s_h'', a_h'')}) }\cdot \left(\sum_{h=0}^H (\phi(s_h^i, a_h^i) - \phi(s_h', a_h'))\right).
\end{align*}
Let $x_{\tau,\tau'}^i = \sum_{h=0}^H (\phi(s_h, a_h) - \phi(s_h', a_h'))$, where $\tau = \{(s_h, a_h)\}_{h\in[H]}$, $\tau' = \{(s_h', a_h')\}_{h\in[H]}$. 
The Hessian of the negative log likelihood can be written as
\begin{align*}
&\nabla^2 \loss_\mathcal{D}(\theta) \\ 
 = &\frac{1}{\numobs} \sum_{i=1}^{\numobs}
\sum_{\tau\in\mathcal{T}(s_0^i)} 
\sum_{\tau'\in\mathcal{T}(s_0^i)} \frac{\exp(\sum_{h=0}^H \inprod{\theta}{\phi(s_h, a_h)+\phi(s_h', a_h')})}{2(\sum_{\tau''\in\mathcal{T}(s_0^i)} \exp(\sum_{h=0}^H \inprod{\theta}{\phi(s_h'', a_h'')}))^2} \cdot x^i_{\tau,\tau'} x_{\tau,\tau'}^{i\top}.
\end{align*}

Since $\exp(\inprod{\theta}{\phi})\in[\exp(-LB), \exp(LB)]$, we know that the coefficients satisfy
\begin{align*}
\frac{\exp(\sum_{h=0}^H \inprod{\theta}{\phi(s_h, a_h)+\phi(s_h', a_h')})}{2(\sum_{\tau''\in\mathcal{T}(s_0^i)} \exp(\sum_{h=0}^H \inprod{\theta}{\phi(s_h'', a_h'')}))^2}  \geq \frac{\exp(-4LB)}{2\sup_s|\mathcal{T}(s)|^2}.
\end{align*}
Set $\gamma = \exp(-4LB)/2$.
We can verify that   for any vector $v\in\mathbb{R}^K$, one has
\begin{align*}
    v^\top \nabla^2 \loss_\mathcal{D}(\theta)v 
 \geq {\gamma}  v^\top \Sigma_{\mathcal{D}} v=  \gamma \|v\|_{\Sigma_{\mathcal{D}}}^2.
\end{align*}
Thus we know that $\loss$ is $\gamma$-strongly convex with respect to the semi-norm $\|\cdot\|_{\Sigma_{\mathcal{D}}}$.

\paragraph{Bounding the estimation error.} Now we aim at bounding the estimation error $\|{\hat\theta_{\mathsf{MLE}} - \theta^\star}\|_{\Sigma_{\mathcal{D}+\lambda I}}$.

Since $\hat\theta_{\mathsf{MLE}}$ is optimal for $\loss_\mathcal{D}$, we have $\loss_\mathcal{D}(\hat\theta_{\mathsf{MLE}}) \leq
\loss_\mathcal{D}(\theta^\star)$.  Defining the error vector $\Delta = \hat\theta_{\mathsf{MLE}} -
\theta^\star$, adding and subtracting the quantity $\inprod{\nabla
  \loss_\mathcal{D}(\theta^\star)}{\Delta}$ yields the bound
\begin{align*}
\loss_\mathcal{D}(\theta^\star + \Delta) - \loss_\mathcal{D}(\theta^\star) - \inprod{\nabla
  \loss_\mathcal{D}(\theta^\star)}{\Delta} & \leq - \inprod{\nabla
  \loss_\mathcal{D}(\theta^\star)}{\Delta}.
\end{align*}
By the $\strongcon$-convexity condition, the left-hand side is lower
bounded by $\strongcon \LAPNORM{\Delta}^2$.  As for the right-hand
side, note that $|\inprod{\nabla
  \loss_\mathcal{D}(\theta^\star)}{\Delta}| \leq \|{\nabla \loss_\mathcal{D}(\theta^\star)}\|_{(\Sigma_{\mathcal{D}}+\lambda I)^{-1}} \;
\|{\Delta}\|_{\Sigma_{\mathcal{D}}+\lambda I}$ for any $\lambda>0$. Altogether we have
\begin{align*}
    \strongcon \LAPNORM{\Delta}^2\leq \|{\nabla \loss_\mathcal{D}(\theta^\star)}\|_{(\Sigma_{\mathcal{D}}+\lambda I)^{-1}} \;
\|{\Delta}\|_{\Sigma_{\mathcal{D}}+\lambda I}. 
\end{align*}

Now we further bound the term $\|{\nabla \loss_\mathcal{D}(\theta^\star)}\|_{(\Sigma_{\mathcal{D}}+\lambda I)^{-1}}$. 
Observe that the gradient takes the form 
\begin{align}\label{eq:gradient_irl}
\nabla \loss_\mathcal{D}(\theta^\star) & 
= -\frac{1}{\numobs} \sum_{i=1}^{\numobs}
\sum_{\tau'\in\mathcal{T}(s_0^i)}  \frac{\exp(\sum_{h=0}^H \inprod{\theta^\star}{\phi(s_h', a_h')})}{\sum_{\tau''\in\mathcal{T}(s_0^i)} \exp(\sum_{h=0}^H \inprod{\theta^\star}{\phi(s_h'', a_h'')}) }\cdot \left(\sum_{h=0}^H (\phi(s_h^i, a_h^i) - \phi(s_h', a_h'))\right).
\end{align}
We set  $X$ as the concatenated   differencing vector 
$x^i_{\tau,\tau'}$ where $\tau,\tau'$ are distinct and ordered.  We also define $V^i_{\tau,\tau'}$ be the random variable of the coefficient of $x^i_{\tau,\tau'}$ in Equation (\ref{eq:gradient_irl}), i.e.
\begin{align*}
    V^i_{\tau,\tau'} = \begin{cases}
\frac{\exp(\sum_{h=0}^H \inprod{\theta^\star}{\phi(s_h', a_h')})}{\sum_{\tau''\in\mathcal{T}(s_0^i)} \exp(\sum_{h=0}^H \inprod{\theta^\star}{\phi(s_h'', a_h'')}) }, & \text{ if }
\tau = \{(s^i_h, a^i_h)\}_{h\in[H]}, \\ 
 -\frac{\exp(\sum_{h=0}^H \inprod{\theta^\star}{\phi(s_h, a_h)})}{\sum_{\tau''\in\mathcal{T}(s_0^i)} \exp(\sum_{h=0}^H \inprod{\theta^\star}{\phi(s_h'', a_h'')}) },  & \text{ if }
\tau' = \{(s^i_h, a^i_h)\}_{h\in[H]}, \\ 
0 & \text{ otherwise. } 
 \end{cases}
\end{align*}
Let $\tilde V_i $ be the concatenated random vector of $\{V^i_{\tau,\tau'}\}$, $V$ be the concatenated random vector of $\{\tilde V_i\}_{i=1}^n$. We know that $\tilde V_i$ and $\tilde V_j$ are independent for each $i\neq j$ due to the independent sampling procedure. We can also verify that the mean of $\tilde V_i$ is $0$. We know that $\tilde V_i$ has almost $\sup_s |\mathcal{T}(s)|$ non-zero elements. And the sum of their absolute value is  bounded by $1$. we know $\tilde V_i$ is $1$-sub-Gaussian.   Now we know that the term  $\|{\nabla \loss_\mathcal{D}(\theta^\star)}\|_{(\Sigma_{\mathcal{D}}+\lambda I)^{-1}}^2$ can be written as
\begin{align*}
\|{\nabla \loss_\mathcal{D}(\theta^\star)}\|_{(\Sigma_{\mathcal{D}}+\lambda I)^{-1}}^2 = \frac{1}{n^2} V^\top 
X  (\Sigma_{\mathcal{D}}+\lambda I)^{-1} X^\top V.
\end{align*}
Let $M = \frac{\sup_s |\mathcal{T}(s)|^2}{n} I$. One can verify that $M \succeq \frac{1}{n^2}X  (\Sigma_{\mathcal{D}}+\lambda I)^{-1} X^\top $ almost surely since $\lambda_{\mathsf{max}}(X  (\Sigma_{\mathcal{D}}+\lambda I)^{-1} X^\top/n^2)\leq \sup_s |\mathcal{T}(s)|^2/ n$. Thus we can upper bound the original term as
\begin{align*}
\|{\nabla \loss_\mathcal{D}(\theta^\star)}\|_{(\Sigma_{\mathcal{D}}+\lambda I)^{-1}}^2 \leq \frac{\sup_s |\mathcal{T}(s)|^2}{n} \|V\|_2^2. 
\end{align*}
By Bernstein's inequality for  sub-Gaussian random variables in quadratic form (see e.g. \citet[Theorem 2.1]{hsu2012tail}), we know that with probability at least $1-\delta$,
\begin{align*}
    \|V\|_2^2\leq  C \cdot (d+\log(1/\delta)).
\end{align*}
Thus altogether, we have \begin{align*}
    \strongcon \LAPNORM{\Delta}^2\leq  \sqrt{ \frac{C \sup_s |\mathcal{T}(s)|^2 \cdot (d+\log(1/\delta))}{n}}
\|{\Delta}\|_{\Sigma_{\mathcal{D}}+\lambda I}. 
\end{align*}
Similar to the pairwise comparison analysis in Appendix~\ref{proof:mle_estimation}, we can derive that with probability at least $1-\delta$,
\begin{align*}
    \|\hat\theta_{\mathsf{MLE}}-\theta^\star\|_{\Sigma_{\mathcal{D}}+\lambda I}\leq C\cdot \sqrt{\frac{\sup_s |\mathcal{T}(s)|^2  (d+\log(1/\delta))}{n} + \lambda B^2}.
\end{align*}

The rest of the proof on the sub-optimality upper bound follows the same argument as Theorem~\ref{thm:LCB_upper_RL_traj}. 

\end{proof}

\subsection{Proof of Theorem~\ref{thm:nonconvex}}\label{proof:nonconvex}
 
To simplify the notation, we 
let $f_\theta^i=r_\theta(s^i, a_1^i) - r_\theta(s^i, a_0^i) $. We can see that   the gradient of $\loss$ takes the form
\begin{align*}
\nabla \loss_\mathcal{D} (\theta) & = \frac{-1}{\numobs }
\sum_{i=1}^{\numobs} \left[ \Ind[\obs^i = 1] \frac{
    \exp(-f_\theta^i)}{1+\exp(-f_\theta^i))} - \Ind[\obs^i=0] \frac{
    1}{1+\exp(-f_\theta^i))}\right] \nabla f_\theta^i.
\end{align*}
And the Hessian of $\ell$ is
\begin{align*}
\nabla^2 \loss_\mathcal{D}(\theta) & =\frac{1}{\numobs}
\sum_{i=1}^{\numobs} \left(\frac{\exp(f_\theta^i)}{(\exp(f_\theta^i)+1)^2}  \cdot \nabla f_\theta^i \nabla f_\theta^{i\top} - \frac{1(y^i=1)\cdot\exp(-f_\theta^i)}{1+\exp(-f_\theta^i)}\cdot \nabla^2 f_\theta^i  + \frac{1(y^i=0)\cdot\exp(f_\theta^i)}{1+\exp(f_\theta^i)}\cdot \nabla^2 f_\theta^i\right). 
\end{align*}
Now from Assumption~\ref{ass:nonconvex}, we have 
\begin{align*}
\nabla^2 \loss_\mathcal{D}(\theta) \succeq \frac{1}{\numobs}
\sum_{i=1}^{\numobs}  \gamma \nabla f_\theta^i \nabla f_\theta^{i\top} -  2\alpha_2 I. 
\end{align*}
where $\strongcon =   \frac{1}{2+\exp(-2LB)+\exp(2LB)}$. Now from the Lipschitz gradient assumption we also know that $\|\nabla f_{\theta}^i - \nabla f_{\theta^\star}^i \|\leq 2\alpha_2\|\theta^\star-\theta\|$. Let $u = \nabla f_{\theta}^i - \nabla f_{\theta^\star}^i$, we have
\begin{align*}
\nabla^2 \loss_\mathcal{D}(\theta) & \succeq \frac{1}{\numobs}
\sum_{i=1}^{\numobs}  \gamma (\nabla f_{\theta^\star}^i+u) (\nabla f_{\theta^\star}^i+u)^{\top} -  2\alpha_2 I \\ 
& \succeq \frac{1}{\numobs}
\sum_{i=1}^{\numobs}  \gamma \nabla f_{\theta^\star}^i\nabla f_{\theta^\star}^{i\top} +  \gamma (\nabla f_{\theta^\star}^i u^\top + u \nabla f_{\theta^\star}^{i\top})-2\alpha_2 I.
\end{align*}
Since $u^\top v\leq \|u\|_2\|v\|_2\leq 2\alpha_2 B\|v\|_2$, $ v^\top\nabla f_{\theta^\star}^i\leq \alpha_1\|v\|_2$,
this gives that
\begin{align*}
 v^\top \nabla^2 \loss_\mathcal{D} (\theta) v \geq \frac{\strongcon}{\numobs
   } \|X v\|_2^2 - 2\alpha_2(1+2\gamma \alpha_1B)\|v\|_2^2 \qquad \mbox{for all $v$},
\end{align*}
where  $\Xmat \in \real^{\numobs \times \numitems}$ has the
 vector $\nabla f_{\theta^\star}^i \in \real^\numitems$ as its $i^{th}$ row.
Thus, if we introduce the error vector $\Delta \coloneqq \hat\theta_{\mathsf{MLE}} -
\theta^\star$, then we may conclude that
\begin{align*}
\loss_\mathcal{D}(\theta^\star + \Delta) - \loss_\mathcal{D}( \theta^\star ) - \inprod{\nabla \loss_\mathcal{D}
  (\theta^\star)}{ \Delta} & \geq \frac{\strongcon}{\numobs }
\|X \Delta\|_2^2  - 2\alpha_2(1+2\gamma \alpha_1B)\|\Delta\|_2^2  =   {\strongcon}
\|{\Delta}\|_{\Sigma_{\mathcal{D}}}^2 - 2\alpha_2(1+2\gamma \alpha_1B)\|\Delta\|_2^2.
\end{align*}

\paragraph{Bounding the estimation error.} Now we aim at bounding the estimation error $\LAPNORM{\hat\theta_{\mathsf{MLE}} - \theta^\star}$. 
Since $\hat\theta_{\mathsf{MLE}}$ is optimal for $\loss_\mathcal{D}$, we have $\loss_\mathcal{D}(\hat\theta_{\mathsf{MLE}}) \leq
\loss_\mathcal{D}(\theta^\star)$.  (When $\hat\theta_{\mathsf{MLE}}$ is approximately optimal, i.e. $\loss_\mathcal{D}(\hat\theta_{\mathsf{MLE}}) \leq \min_\theta
\loss_\mathcal{D}(\theta)+\epsilon$, the same argument also holds up to an extra additive term $\epsilon$.) Defining the error vector $\Delta = \hat\theta_{\mathsf{MLE}} -
\theta^\star$, adding and subtracting the quantity $\inprod{\nabla
  \loss_\mathcal{D}(\theta^\star)}{\Delta}$ yields the bound
\begin{align*}
\loss_\mathcal{D}(\theta^\star + \Delta) - \loss_\mathcal{D}(\theta^\star) - \inprod{\nabla
  \loss_\mathcal{D}(\theta^\star)}{\Delta} & \leq - \inprod{\nabla
  \loss_\mathcal{D}(\theta^\star)}{\Delta}.
\end{align*}
We know the left-hand side is lower
bounded by $\strongcon \LAPNORM{\Delta}^2-2\alpha_2(1+2\gamma \alpha_1B)\|\Delta\|_2^2$.  As for the right-hand
side, note that $|\inprod{\nabla
  \loss_\mathcal{D}(\theta^\star)}{\Delta}| \leq \|{\nabla \loss_\mathcal{D}(\theta^\star)}\|_{(\Sigma_{\mathcal{D}}+\lambda I)^{-1}} \;
\|{\Delta}\|_{\Sigma_{\mathcal{D}}+\lambda I}$ for any $\lambda>0$. Altogether we have
\begin{align*}
    \strongcon \LAPNORM{\Delta}^2\leq \|{\nabla \loss_\mathcal{D}(\theta^\star)}\|_{(\Sigma_{\mathcal{D}}+\lambda I)^{-1}} \;
\|{\Delta}\|_{\Sigma_{\mathcal{D}}+\lambda I} + \beta \|\Delta\|_2^2,
\end{align*}

where $\beta = 2\alpha_2(1+2\gamma \alpha_1 B)$. 
Now we further bound the term $\|{\nabla \loss_\mathcal{D}(\theta^\star)}\|_{(\Sigma_{\mathcal{D}}+\lambda I)^{-1}}$. 
Observe that the gradient takes the form
\begin{align*}
\nabla \loss_\mathcal{D} (\theta^\star) & = \frac{-1}{\numobs }
\sum_{i=1}^{\numobs} \left[ \Ind[\obs^i = 1] \frac{
    \exp(-f_{\theta^\star}^i)}{1+\exp(-f_{\theta^\star}^i))} - \Ind[\obs^i=0] \frac{
    1}{1+\exp(-f_{\theta^\star}^i))}\right] \nabla f_{\theta^\star}^i.
\end{align*}
Define a random vector $V \in \real^\numobs$ with independent
components as
\begin{align*}
V_i = \begin{cases} \frac{
    \exp(-f_{\theta^\star}^i)}{1+\exp(-f_{\theta^\star}^i))} &
  \mbox{w.p. \quad}\frac{
    1}{1+\exp(-f_{\theta^\star}^i))}\\
\frac{
    -1}{1+\exp(-f_{\theta^\star}^i))}& \mbox{w.p. \quad} \frac{
    \exp(-f_{\theta^\star}^i)}{1+\exp(-f_{\theta^\star}^i))}.
\end{cases}
\end{align*}
With this notation, we have $\nabla \loss_\mathcal{D}(\theta^\star) = -
\frac{1}{\numobs } \, \Xmat^\top V$. One can verify that
$\Exs[V] = 0$ and $| V_i| \leq 1$. 

Defining the $\numobs$-dimensional square matrix $M \coloneqq
\frac{1}{ \numobs^2} \diffmx (\Sigma_{\mathcal{D}}+\lambda I)^{-1} \diffmx^\top$, we have $\|{\nabla \loss_\mathcal{D}(\theta^\star)}\|_{(\Sigma_{\mathcal{D}}+\lambda I)^{-1}}
=V^\top M V$.
Let the eigenvalue decomposition of $X^\top X$ be $X^\top X= U \Lambda U^\top$. We can bound the trace and operator norm of $M$ as 
\begin{align*}
\mathsf{Tr}(M) &= \frac{1}{n^2}\mathsf{Tr}(U(\Lambda/n+\lambda I)^{-1} U^\top U\Lambda U^\top) \leq  \frac{d}{
  \numobs}  \\
  \mathsf{Tr}(M^2) &= \frac{1}{n^4}\mathsf{Tr}(U(\Lambda/n+\lambda I)^{-1} U^\top U\Lambda U^\top U(\Lambda/n+\lambda I)^{-1} U^\top U\Lambda U^\top) \leq  \frac{d}{
  \numobs^2}  \\
  \|M\|_{\mathsf{op}} &= \lambda_{\mathsf{max}}(M)\leq 
\frac{1}{ \numobs},
\end{align*} 
Moreover, since the components of $V$ are independent and of zero mean, and $|V_i| \leq 1$, the variables are
$1$-sub-Gaussian, and hence the Bernstein's inequality for  sub-Gaussian random variables in quadratic form (see e.g. \citet[Theorem 2.1]{hsu2012tail})
implies that  with probability at least $1-\delta$, 
\begin{align*}
    \|{\nabla \loss_\mathcal{D}(\theta^\star)}\|^2_{(\Sigma_{\mathcal{D}}+\lambda I)^{-1}} =V^\top M V \leq C_1\cdot \frac{d+\log(1/\delta)}{n}.
\end{align*}
Here $C_1$ is some universal constant. This gives us 
\begin{align*}
    \strongcon \|{\Delta}\|_{\Sigma_{\mathcal{D}}+\lambda I}^2&\leq \|{\nabla \loss_\mathcal{D}(\theta^\star)}\|_{(\Sigma_{\mathcal{D}}+\lambda I)^{-1}} \;
\|{\Delta}\|_{\Sigma_{\mathcal{D}}+\lambda I} +4(\lambda\gamma+2\alpha_2(1+2\gamma \alpha_1B)) B^2\\ 
& \leq \sqrt{C_1\cdot \frac{d+\log(1/\delta)}{ n}}\|{\Delta}\|_{\Sigma_{\mathcal{D}}+\lambda I} +4(\lambda\gamma+2\alpha_2(1+2\gamma \alpha_1B)) B^2.
\end{align*}
Solving the above inequality gives us that for some constant $C_2$,
\begin{align*}
    \|{\Delta}\|_{\Sigma_{\mathcal{D}}+\lambda I}\leq C_2\cdot \sqrt{ \frac{d+\log(1/\delta)}{\gamma^2 n} +  (\lambda+\alpha_2/\gamma+\alpha_1\alpha_2 B) B^2}.
\end{align*}